\documentclass[]{IOS-Book-Article}
\usepackage{nesybook}
\usepackage{chap}

\begin{document}

% \pagestyle{headings}
% \def\thepage{}

%\maketitle

\setcounter{tocdepth}{0}

%\tableofcontents
%\setcounter{page}{0}
%\pagestyle{headings}

\chapter{Combining Probabilistic Logic and Deep Learning for Self-Supervised Learning}
\label{Hoifung_Probabilistic_logic:chap}
\chapterauthor{Hoifung Poon}{Microsoft Research, Redmond WA, USA}
\chapterauthor{Hai Wang}{JD Finance America Corporation, Mountain View CA, USA}
\chapterauthor{Hunter Lang}{MIT CSAIL, Cambridge MA, USA}
\allchapterauthors{Hoifung Poon, Hai Wang, Hunter Lang}

\setlength{\floatsep}{5pt}
\setlength{\textfloatsep}{5pt}

{\footnotesize
Deep learning has proven effective for various application tasks, but its applicability is limited by the reliance on annotated examples. Self-supervised learning has emerged as a promising direction to alleviate the supervision bottleneck, but existing work focuses on leveraging co-occurrences in unlabeled data for task-agnostic representation learning, as exemplified by masked language model pretraining. In this chapter, we explore {\em task-specific self-supervision}, which leverages domain knowledge to automatically annotate noisy training examples for end applications, either by introducing labeling functions for annotating individual instances, or by imposing constraints over interdependent label decisions. We first present {\em deep probabilistic logic} (DPL), which offers a unifying framework for task-specific self-supervision by composing probabilistic logic with deep learning. DPL represents unknown labels as latent variables and incorporates diverse self-supervision using probabilistic logic to train a deep neural network end-to-end using variational EM. Next, we present {\em self-supervised self-supervision} (S4), which adds to DPL the capability to learn new self-supervision automatically. Starting from an initial seed self-supervision, S4 iteratively uses the deep neural network to propose new self supervision. These are either added directly (a form of {\em structured self-training}) or verified by a human expert (as in {\em feature-based active learning}). Experiments on real-world applications such as biomedical machine reading and various text classification tasks show that task-specific self-supervision can effectively leverage domain expertise and often match the accuracy of supervised methods with a tiny fraction of human effort.
}    

\smallsection{Introduction}
\vspace{-5pt}
\begin{figure}
    \centering
    \includegraphics[width=0.8\linewidth]{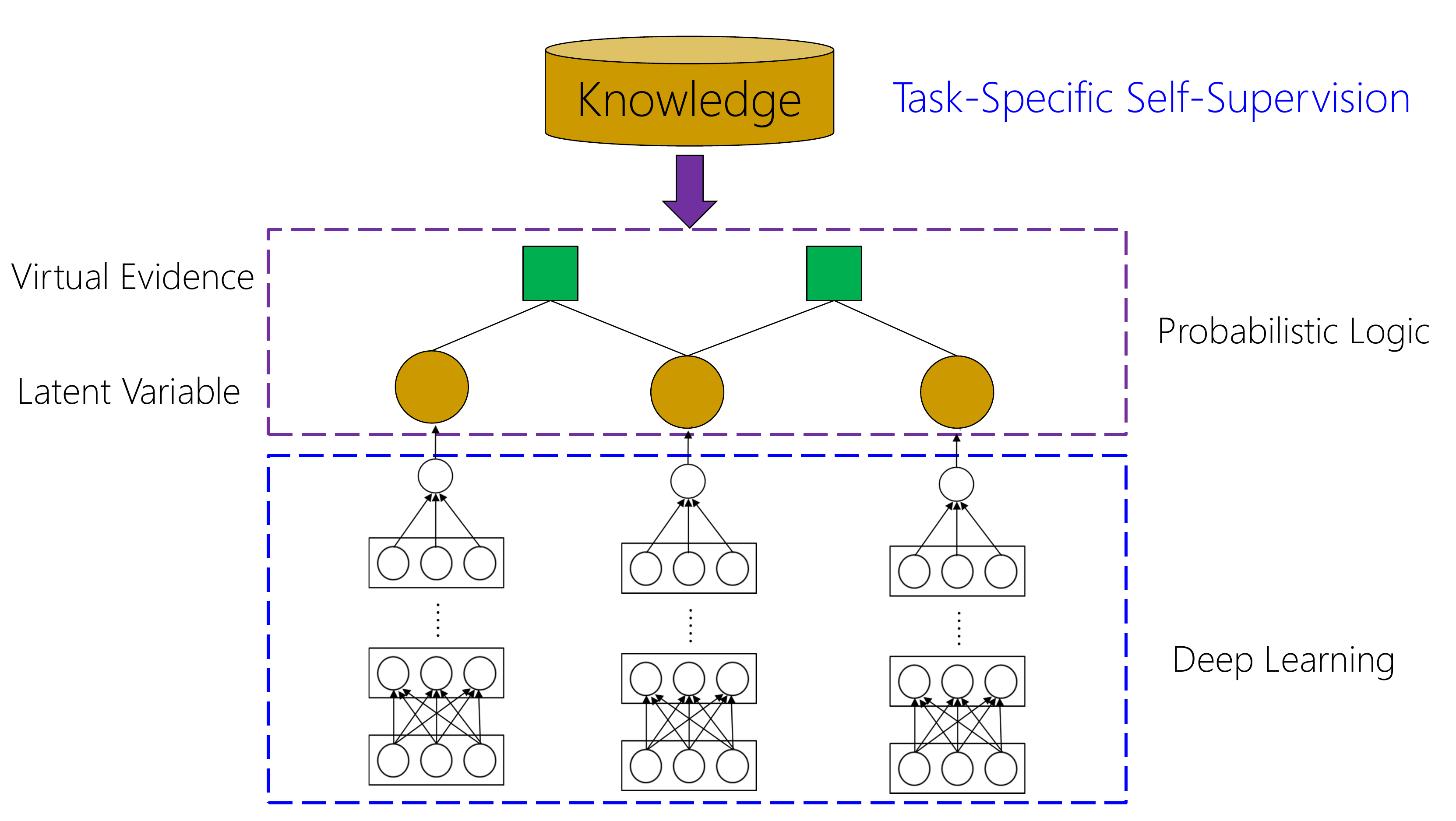}
    \vspace{-5pt}
    \caption{Deep Probabilistic Logic: A general framework for combining task-specific self-supervision strategies by composing probabilistic logic with deep learning. Learning amounts to maximizing conditional likelihood of virtual evidence given input by summing up latent label decisions.
    }
    \label{fig:DPL}
\end{figure}

Machine learning has made great strides in enhancing model sophistication and learning efficacy, as exemplified by recent advances in deep learning \cite{lecun2015deep}. However, contemporary supervised learning techniques require a large amount of labeled data, which is expensive and time-consuming to produce. This problem is particularly acute in specialized domains like biomedicine, where crowdsourcing is difficult to apply.
Self-supervised learning has emerged as a promising paradigm to overcome the annotation bottleneck by automatically generating noisy training examples from unlabeled data. 
However, existing work predominantly focuses on {\em task-agnostic representation learning} by modeling co-occurrences in unlabeled data, as exemplified in masked language model pretraining \cite{devlin2018bert}. 
While successful, it still requires manual annotation of task-specific labeled examples to fine-tune pretrained models for end applications.

In this chapter, we explore {\em task-specific self-supervised learning}, which can directly generate training examples for end applications. The key insight is to {\em introduce a general framework that can best leverage human expert bandwidth by accepting and leveraging alternative forms of supervision}. For example, with relatively small effort, domain experts can produce labeling functions \cite{ratner2016data,bach&al17} and joint inference constraints \cite{chang&al07,poon&domingos08,druck&mccallum08,ganchev&al10}. Additionally, there are ample available resources such as ontologies and knowledge bases that can be leveraged to automatically annotate training examples in unlabeled data \cite{craven1999constructing,mintz2009distant}. 
Using such self-specified supervision templates, we can automatically produce training examples at scale from unlabeled data.

A central challenge in leveraging such diverse forms of self-supervision is that they are inherently noisy and may be contradictory with each other. We present deep probabilistic logic (DPL), which provides a unifying framework for task-specific self-supervision by composing probabilistic logic with deep learning \cite{wang2018deep}. 
DPL models label decisions as latent variables, represents prior knowledge on their relations using Markov logic (weighted first-order logical formulas) \cite{richardson&domingos06}, and alternates between learning a deep neural network for the end task and refining uncertain formula weights for self-supervision, using variational EM. 
Probabilistic logic offers a principled way to combine noisy and inconsistent self-supervision. Feedback from deep learning helps resolve noise and inconsistency in the initial self-supervision. Experiments on biomedical machine reading show that distant supervision, data programming, and joint inference can be seamlessly combined in DPL to substantially improve machine reading accuracy, without requiring any manually labeled examples.

\setlength{\textfloatsep}{10pt}
\begin{figure*}[t]
\centering
\includegraphics[width=0.9\linewidth]{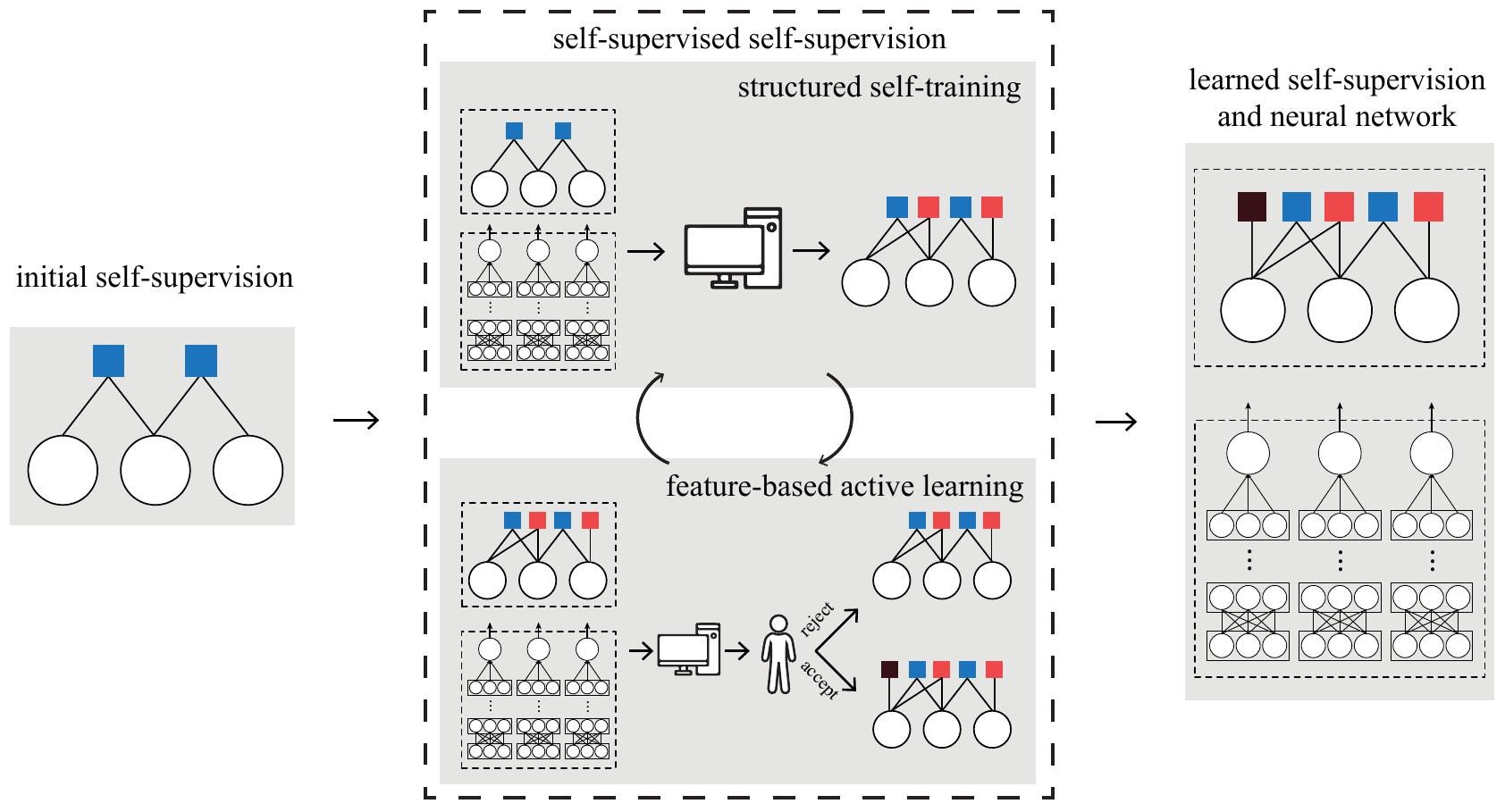}
\caption{Self-Supervised Self-Supervision (S4): S4 builds on deep probabilistic logic and uses probabilistic logic to represent self-supervision for learning a deep neural network for the end prediction task. 
Starting from pre-specified self-supervision, S4 interleaves structure learning and active learning steps to introduce new self-supervision for training the neural network and refining the graphical model parameters. Self-supervision factors from initialization, structure learning, and active learning are shown in blue, red, and black, respectively. 
}
\label{fig:S4}
\end{figure*}

DPL still requires human experts to manually specify self-supervision. 
We next present {\em Self-Supervised Self-Supervision (S4)}, which is a general framework for learning to add new self-supervision, by extending DPL with structure learning and active learning capabilities (see Figure~\ref{fig:S4}) \cite{lang&poon21}. S4 first runs DPL using the pre-specified seed self-supervision (represented as a Markov logic network), then iteratively proposes new self-supervision (weighted logical formula) using the trained deep neural network, and determines whether to add it directly to the self-supervision Markov logic network or ask a human expert to vet it. The former can be viewed as {\em structured self-training}, which generalizes self-training \cite{mcclosky2006effective} by adding not only individual labels but also arbitrary probabilistic factors over them.
The latter subsumes {\em feature-based active learning} \cite{druck2009active} with arbitrary features expressible using probabilistic logic. 
By combining the two in a unified framework, S4 can leverage both paradigms for generating new self-supervision and subsume many related approaches as special cases. 
Using transformer-based models \cite{vaswani2017attention} for the deep neural network in DPL, we explore various self-supervision proposal mechanisms based on neural attention and label entropy. 
Our method can learn to propose both unary potential factors over individual labels and joint-inference factors over multiple labels.
Experiments on various text classification tasks show that S4 can substantially improve over the seed self-supervision by proposing new self-supervision, often matching the accuracy of fully supervised systems with a fraction of human effort.

We first review related work in self-supervised learning and combining probabilistic logic with deep learning. We then present deep probabilistic logic and self-supervised self-supervision in details. We conclude by discussing several exciting future directions. We focus on natural language processing (NLP) applications in this chapter, but our methods are potentially applicable to other domains and tasks.

\smallsection{Task-Specific Self-Supervision}
\vspace{-5pt}
Techniques to compensate for the lack of direct supervision come in many names and forms \cite{mintz2009distant, ratner2016data,bach2017learning,roth2017incidental,wang2018deep}.
Self-supervision has emerged as an encompassing paradigm that views these as instances of using self-specified templates to generate noisy labeled examples on unlabeled data.
The name {\em self-supervision} is closely related to {\em self-training} \cite{mcclosky2006effective}, which bootstraps from a supervised classifier, uses it to annotate unlabeled instances, and iteratively adds confident labels to retrain the classifier. 
{\em Task-agnostic self-supervision} generalizes word embedding and language modeling by learning to predict self-specified masked tokens, as exemplified by recent pretraining methods such as BERT \cite{devlin2018bert}.
In this chapter, we focus on {\em task-specific self-supervision} and use pretrained models as a building block for task-specific learning. Below, we review three popular forms of task-specific self-supervision: distant supervision, data programming, and joint inference.

\smallpar{Distant supervision}
This paradigm was first introduced for binary relation extraction \cite{craven1999constructing,mintz2009distant}. 
In its simplest form, distant supervision generates a positive example if an entity pair with a known relation co-occurs in a sentence, and samples negative examples from co-occurring entity pairs not known to have the given relation.
It has recently been extended to cross-sentence relation extraction \cite{quirkpoon2017,peng&al17}. In principle, one simply looks beyond single sentences for co-occurring entity pairs. However, this can introduce many false positives and prior work used a small sliding window and filtering (minimal-span) to mitigate training noise. Even so, accuracy is relatively low. 
Both \cite{quirkpoon2017} and \cite{peng&al17} used ontology-based string matching for entity linking, which also incurs many false positives, as entity mentions can be highly ambiguous (e.g., PDF and AAAS are gene names). 
Distant supervision for entity linking is relatively underexplored, and prior work generally focuses on Freebase entities, where links to the corresponding Wikipedia articles are available for learning \cite{huang2015leveraging}.

\smallpar{Data Programming} 
Instead of annotated examples, domain experts are asked to produce labeling functions, each of which assigns a label to an instance if the input satisfies certain conditions, often specified by simple rules \cite{ratner&al16}. 
This paradigm is useful for semantic tasks, as high-precision text-based rules are often easy to come by. 
However, there is no guarantee on broad coverage, and labeling functions are still noisy and may contradict with each other.
The common denoising strategy assumes that labeling functions make random mistakes, and focuses on estimating their accuracy and correlation \cite{ratner2016data, bach2017learning,varma2017inferring}.
A more sophisticated strategy also models instance-level labels and uses instance embedding to estimate instance-level weight for each labeling function \cite{liu2017heterogeneous}.

\smallpar{Joint Inference}
Distant supervision and data programming focus on infusing task-specific self-supervision on individual labels. 
Additionally, there is rich linguistic and domain knowledge that does not specify values for individual labels, but imposes constraints on their joint distribution.
For example, if two mentions are coreferent, they should agree on entity properties \cite{poon&domingos08}.
There is a rich literature on joint inference for NLP applications. Notable methodologies include constraint-driven learning \cite{chang&al07}, general expectation \cite{druck&mccallum08}, posterior regularization \cite{ganchev&al10}, and probabilistic logic \cite{poon&domingos08}.
Constraints can be imposed on relational instances or on model expectations.
Learning and inference are often tailor-made for each approach, including beam search, primal-dual optimization, weighted satisfiability solvers, etc.
Recently, joint inference has also been used in denoising distant supervision. 
Instead of labeling all co-occurrences of an entity pair with a known relation as positive examples, one only assumes that at least one instance is positive \cite{MultiR,lin&al16}. 

Existing self-supervision paradigms are typically special cases of deep probabilistic logic (DPL). E.g., data programming admit only self-supervision for individual instances (labeling functions or their correlations). 
{\em Anchor learning} \cite{halpern2016electronic} is an earlier form of data programming that, while more restricted, allows for stronger theoretical learning guarantees. 
{\em Prototype learning} is an even earlier special case with labeling functions provided by ``prototypes'' \cite{haghighi2006prototype, poon2013grounded}.
Using Markov logic to model self-supervision, DPL can incorporate arbitrary prior beliefs on both individual labels and their interdependencies, thereby unleashing the full power of {\em joint inference} \cite{chang2007guiding, druck2008learning, poon2008joint, ganchev2010posterior} to amplify and propagate self-supervision signals.

Self-supervised self-supervision (S4) further extends DPL with structure learning capability. Most structure learning techniques are developed for the supervised setting, where structure search is guided by labeled examples \cite{koller-struc-lrn,kok2005learning}. 
Moreover, traditional relational learning induces deterministic rules and is susceptible to noise and uncertainty.
{\em Bootstrapping} learning is one of the earliest and simplest self-supervision methods with some rule-learning capability, by alternating between inducing characteristic contextual patterns and classifying instances \cite{hearst1992automatic, NELL}. The pattern classes are limited and only applicable to special problems (e.g., ``A such as B" to find $\tt ISA$ relations). Most importantly, they lack a coherent probabilistic formulation and may suffer catastrophic {\em semantic drift} due to ambiguous patterns (e.g., ``cookie'' as food or compute use). 
\cite{yarowsky1995unsupervised} and \cite{collins1999unsupervised} designed a more sophisticated rule induction approach, but their method uses deterministic rules and may be sensitive to noise and ambiguity.
Recently, Snuba \cite{varma2018snuba} extends the data programming framework by automatically adding new labeling functions, but like prior data programming methods, their self-supervision framework is limited to modeling prior beliefs on individual instances. Their method also requires access to a small number of labeled examples to score new labeling functions.

Another significant advance in S4 is by extending DPL with the capability to conduct {\em structured active learning}, where human experts are asked to verify arbitrary virtual evidences, rather than a label decision.
Note that by admitting joint inference factors, this is more general than prior use of {\em feature-based active learning}, which focuses on per-instance features \cite{druck2009active}.
As our experiments show, interleaving structured self-training learning and structured active learning results in substantial gains, and provides the best use of precious human bandwidth. \cite{tong2001active} previously considered active structure learning in the context of Bayesian networks.
Anchor learning \cite{halpern2016electronic} can also suggest new self-supervision for human review.
Darwin \cite{galhotra2020adaptive} incorporates active learning for verifying proposed rules, but it doesn't conduct structure learning, and like Snuba and other data programming methods, it only models individual instances.

\smallsection{Combining Probabilistic Logic with Deep Learning}
\vspace{-5pt}
Probabilistic logic combines logic's expressive power with the capability of graphical models to handle uncertainty. 
A representative example is Markov logic \cite{richardson&domingos06}, which defines a probability distribution using weighted first-order logical formulas as templates for a Markov model. 
Given random variables $X$ for a problem domain, Markov logic uses a set of weighted first-order logical formulas $(w_i, f_i): i$ to define a joint probability distribution $P(X)\propto \exp\sum_i~w_i\cdot f_i(X)$. 
Intuitively, the weight $w_i$ correlates with how likely formula $f_i$ holds true. 
A logical constraint is a special case when $w_i$ is set to $\infty$ (or a sufficiently large number). 
Probabilistic logic has been applied to incorporating joint inference and other task-specific self-supervision for various NLP tasks \cite{poon&domingos07,poon&domingos08,poon&vanderwende10}, but its expressive power comes at a price: learning and inference are generally intractable, and end-to-end modeling often requires heavy approximation \cite{kimmig2012short}. 

Recently, there has been increasing interest in combining probabilistic logic with deep learning, thereby harnessing probabilistic logic's capability in representing and reasoning with knowledge, as well as leveraging deep learning's strength in distilling complex patterns from high-dimension data. 
Most existing work focuses on incorporating probabilistic logic in inference, typically by replacing discrete logical predicates with neural representation and jointly learning global probabilistic logic parametrization and local neural networks, as exemplified by differentiable proving \cite{rocktaschel2017end} and DeepProbLog \cite{manhaeve2018deepproblog}. 
By contrast, we leverage probabilistic logic to provide self-supervision for deep learning.  These two approaches are complementary and can be combined, i.e., by using probabilistic logic to infuse prior knowledge for both learning and inference, as explored in recent work on knowledge graph embedding \cite{qu&tang19}.

Deep generative models also combine deep learning with probabilistic models, but focus on uncovering latent factors to support generative modeling and semi-supervised learning \cite{kingma&welling13,kingma&al14}.
Knowledge infusion is limited to introducing structures among latent variables (e.g., Markov chain)
\cite{johnson&al16}.
Deep probabilistic programming provides a flexible interface for exploring such composition \cite{edward}.
In deep probabilistic logic and self-supervised self-supervision, we instead combine a discriminative neural network predictor with a generative self-supervision model based on Markov logic, and can fully leverage their respective capabilities to advance co-learning \cite{blum1998combining,grechkin2017ezlearn}. 
Deep neural networks also provide a powerful feature-induction engine to support structure learning and active learning.

\smallsection{Deep Probabilistic Logic: A Unifying Framework for Self-Supervision}
\label{sec:mdl}
\vspace{-5pt}
In this section, we introduce deep probabilistic logic (DPL) as a unifying framework for task-specific self-supervision.
Label decisions are modeled as latent variables. Self-supervision is represented as generalized virtual evidence, and learning maximizes the conditional likelihood of virtual evidence given input. 
The use of probabilistic logic is limited to modeling self-supervision, leaving end-to-end modeling to the deep neural network for end prediction. 
This alleviates the computational challenges in using probabilistic logic end-to-end, while leveraging the strength of deep learning in distilling complex patterns from high-dimension data.

Infusing knowledge in neural network training is a long-standing challenge in deep learning \cite{towell1994knowledge}. 
\cite{hu2016harnessing,hu2016deep} first used logical rules to help train a convolutional neural network for sentiment analysis. DPL draws inspiration from their approach, but is more general and theoretically well-founded. \cite{hu2016harnessing,hu2016deep} focused on supervised learning and the logical rules were introduced to augment labeled examples via posterior regularization \cite{ganchev&al10}. DPL can incorporate both direct and indirect supervision, including posterior regularization and other forms of self-supervision.
Like DPL, \cite{hu2016deep} also refined uncertain weights of logical rules, but they did it in a heuristic way by appealing to symmetry with standard posterior regularization. We provide a novel problem formulation using generalized virtual evidence, which shows that their heuristic is a special case of variational EM and opens up opportunities for other optimization strategies.

We first review the idea of virtual evidence and show how it can be generalized to represent any form of self-supervision. We then formulate the learning objective and show how it can be optimized using variational EM.

\smallsubsection{Deep Probabilistic Logic}
\vspace{-10pt}
Given a prediction task, let $\mathcal{X}$ denote the set of possible inputs and $\mathcal{Y}$ the set of possible outputs. The goal is to train a prediction module $\Psi(x,y)$ that scores output $y$ given input $x$. We assume that $\Psi(x,y)$ defines the conditional probability $P(y|x)$ using a deep neural network with a softmax layer at the top.
Let $X=(X_1,\cdots,X_N)$ denote a sequence of inputs and $Y=(Y_1,\cdots,Y_N)$ the corresponding outputs. We consider the setting where $Y$ are unobserved, and $\Psi(x,y)$ is learned using task-specific self-supervision.

\smallpar{Virtual evidence} 
Pearl \cite{pearl2014probabilistic} first introduced the notion of virtual evidence, which has been used to incorporate label preference in semi-supervised learning \cite{reynolds&bilmes05,subramanya&bilmes07,xiao09} and grounded learning \cite{parikh&al15}.
Suppose we have a prior belief on the value of $y$.
It can be represented by introducing a binary variable $v$ as a dependent of $y$ such that $P(v=1|y=l)$ is proportional to the prior belief of $y=l$.
$v=1$ is thus an observed evidence that imposes soft constraints over $y$. 
Direct supervision (i.e., observing the true label) for $y$ is a special case when the belief is concentrated on a specific value $y=l^*$ (i.e., $P(v=1|y=l)=0$ for any $l\ne l^*$). 
The virtual evidence $v$ can be viewed as a reified variable for a potential function $\Phi(y)\propto P(v=1|y)$. 
This enables us to generalize virtual evidence to arbitrary potential functions $\Phi(X,Y)$.
In the rest of the paper, we will simply refer to the potential functions as virtual evidences, without introducing the reified variables explicitly.

\smallpar{DPL}
Let $K=(\Phi_1,\cdots,\Phi_V)$ be a set of virtual evidence derived from prior knowledge.
DPL comprises of a supervision module over K and a prediction module over all input-output pairs (Figure~\ref{fig:DPL}), and defines a probability distribution:
\vspace{-5pt}
\[P(K,Y|X)\propto \prod_v~\Phi_{v}(X, Y)\cdot\prod_i~\Psi(X_i, Y_i)\]

\vspace{-10pt}
Without loss of generality, we assume that virtual evidences are log-linear factors, which can be compactly represented by weighted first-order logical formulas \cite{richardson&domingos06}. Namely, $\Phi_v(X,Y)=\exp(w_v\cdot f_v(X,Y))$, where $f_v(X,Y)$ is a binary feature represented by a first-order logical formula. A hard constraint is the special case when $w_v=\infty$ (in practice, it suffices to set it to a large number, e.g., 10).
In prior use of virtual evidence, $w_v$'s are generally pre-determined from prior knowledge. However, this may be suboptimal. Therefore, we consider a general Bayesian learning setting where each $w_v$ is drawn from a pre-specified prior distribution $w_v\sim P(w_v|\alpha_v)$. Fixed $w_v$ amounts to the special case when the prior is concentrated on the preset value. For uncertain $w_v$'s, we can compute their maximum a posteriori (MAP) estimates and/or quantify the uncertainty.

\smallpar{Distant supervision}
Virtual evidence for distant supervision is similar to that for direct supervision. For example, for relation extraction, distant supervision from a knowledge base of known relations will set $f_{KB}(X_i,Y_i)=\mathbb{I}[\text{\tt In-KB}(X_i,r) \land Y_i=r]$, where $\text{\tt In-KB}(X_i,r)$ is true iff the entity tuple in $X_i$ is known to have relation $r$ in the KB.

\smallpar{Data programming}
Virtual evidence for data programming is similar to that for distant supervision:
$f_{L}(X_i,Y_i)=\mathbb{I}[L(X_i) = Y_i]$, where $L(X_i)$ is a labeling function provided by domain experts.
Labeling functions are usually high-precision rules, but errors are still common, and different functions may assign conflicting labels to an instance.
Existing denoising strategies either assume that each function makes random errors independently, and resolves the conflicts by weighted votes \cite{ratner&al16}, or attempts to learn and account for the dependencies between labeling functions \cite{ratner2019training}.
In DPL, the former can be done by simply treating error probabilities as uncertain parameters and inferring them during learning, and the latter can be done with structure learning over the virtual evidences. 

\smallpar{Joint inference}
Constraints on instances or model expectations can be imposed by introducing the corresponding virtual evidence \cite{ganchev&al10} (Proposition 2.1). The weights can be set heuristically \cite{chang&al07,mann&mccallum08,poon&domingos08} or iteratively via primal-dual methods \cite{ganchev&al10}.
In addition to instance-level constraints, DPL can incorporate arbitrary high-order soft and hard constraints that capture the interdependencies among multiple instances.
For example, identical mentions in proximity probably refer to the same entity, which is useful for resolving ambiguous mentions by leveraging their unambiguous coreferences (e.g., an acronym in apposition of the full name).
This can be represented by the virtual evidence $f_{\tt Joint}(X_i,Y_i,X_j,Y_j)=\mathbb{I}[{\tt Coref}(X_i,X_j) \land Y_i=Y_j]$, where ${\tt Coref}(X_i,X_j)$ is true iff $X_i$ and $X_j$ are coreferences.
Similarly, the common denoising strategy for distant supervision replaces the mention-level constraints with type-level constraints \cite{MultiR}. %,lin&al16}.
Suppose that $X_E\subset X$ contains all $X_i$'s with co-occurring entity tuple $E$. The new constraints simply impose that, for each $E$ with known relation $r\in KB$, $Y_i=r$ for at least one $X_i\in X_E$. 
This can be represented by a high-order factor on $(X_i,Y_i: X_i\in X_E)$.

\begin{algorithm}[t]
\footnotesize
\begin{algorithmic}
\caption{DPL Learning}\label{alg:learn}
\State \textbf{Input:} Virtual evidences $K=\Phi_{1:V}$, deep neural network $\Psi$, inputs $X=(X_1,\cdots,X_N)$, unobserved outputs $Y=(Y_1,\cdots,Y_N)$.
\State \textbf{Output:} Learned prediction module $\Psi^*$ 
\State \textbf{Initialize:} $\Phi^0 \sim \text{priors}$, $\Psi^0 \sim \text{uniform}$.

\For{\texttt{$t=1:T$}}{
\small{
    \begin{nospaceflalign*}
    %\begin{split}
    %\begin{flalign*}
      &\qquad\qquad q^t(Y) \leftarrow P(Y|\Phi^{t-1}, \Psi^{t-1}, X) \text{ (approximately)} &\\
    %\end{split}
      &\qquad\qquad \Phi^t \leftarrow \arg\min_{\Phi}~D_{KL}(q^t(Y)~||~ \prod_v~\Phi_v(X,Y)) &\\
      &\qquad\qquad \Psi^t \leftarrow  \arg\min_{\Psi}~D_{KL}(q^t(Y)~||~\prod_i~\Psi(X_i,Y_i))&
    \end{nospaceflalign*}
    %\end{flalign*}
    }
}
\State \Return $\Psi^*=\Psi^T$.
\end{algorithmic}
\end{algorithm}  

\smallpar{Parameter learning}
Learning in DPL maximizes the conditional likelihood of virtual evidences $P(K|X)$.
We could directly optimize this objective by summing out the latent variable $Y$ to compute the gradient and run backpropagation.
Instead, in this work we opted for a modular approach using (variational) EM, since both the E- and M-steps reduce to standard inference and learning problems, and because summing out $Y$ is often computationally intractable when using joint inference.
See Algorithm~\ref{alg:learn}.

In the E-step, we compute an approximation $q(Y)$ for $P(Y|K,X)$.
Exact inference is generally intractable, but there is a plethora of approximate inference methods that can efficiently produce an estimate for $P(Y|K,X)$.
In this work we chose to use loopy belief propagation (BP) \cite{murphy1999loopy}.
For each virtual evidence factor $f$, BP computes a ``pseudo-marginal'' $q_f(Y)$, which is an estimate of $\mathbb{E}_{P(Y|K,X)}[f(X,Y)]$.
In particular, BP also computes estimates $q_{BP}(Y_i) \approx \mathbb{E}_{P(Y|K,X)}[Y_i]$, the posterior marginal for each instance's label variable.
We set $q(Y) = \prod_i q_{BP}(Y_i)$, the product of approximate marginals returned by BP.
While this is not theoretically principled (e.g., mean-field iteration may compute a better factored approximation $q(Y)$), BP provides a good balance of speed and flexibility in dealing with complex, higher-order factors.

Note that inference with \emph{high-order factors} of large size (i.e., factors involving many label variables $Y_i$) can still be challenging with BP.
However, there is a rich body of literature for handling such structured factors in a principled way. 
For example, in our biomedical machine reading application, we alter the BP message passing schedule so that each at-least-one factor (for denoising distant supervision; see next subsection) will compute messages to its variables jointly by renormalizing their current marginal probabilities with noisy-or \cite{keith2017identifying}, which is a soft version of dual decomposition \cite{caroe1999dual}.

In the M-step, we treat the approximation $q(Y)$ as giving probabilistic label assignments to $Y$, and use those assignments to optimize $\Phi$ and $\Psi$ via standard supervised learning.
The M-step naturally decouples into two separate optimization tasks, one for the supervision module $\Phi$ and one for the prediction module $\Psi$.
For the prediction module (represented by a deep neural network), the optimization task $\arg\min_{\Psi}~D_{KL}(q^t(Y)~||~\prod_i~\Psi(X_i,Y_i))$ reduces to standard deep learning with cross-entropy loss, where the label distribution over $Y$ is given by our current estimate $q^t(Y_i)$ for the marginal $\mathbb{E}_{P(Y|K,X)}[Y_i]$.
That is, we train the deep network $\Psi$ to minimize the cross-entropy $-\frac{1}{N}\sum_{i}\sum_y q^t(Y_i=y)\log \Psi(X_i, y)$.

For the supervision module $\Phi$, the optimization task $\arg\min_{\Phi}~D_{KL}(q^t(Y)~||~\prod_v~\Phi_v(X,Y))$ similarly reduces to standard parameter learning for log-linear models (i.e., learning all $w_v$'s that are not fixed). Given the estimate for $P(Y|K,X)$ produced in the E-step, this is a convex optimization problem with a unique (under certain conditions) global optimum. 
Here, we simply use gradient descent, with the partial derivative for $w_v$ being
$\mathbb{E}_{\Phi(Y,X)}~[f_v(X,Y)] - \mathbb{E}_{q(Y)}~[f_v(X,Y)]$.
For a tied weight, the partial derivative will sum over all features that originate from the same template.
The second expectation is easy to compute given $q(Y)$ estimated in the E-step.
The first expectation, on the other hand, requires probabilistic inference in the graphical model \emph{excluding} the prediction module $\Psi$. But this can again be computed using belief propagation, similar to the E-step, except that the messages are limited to factors within the supervision module (i.e., messages from $\Psi$ are no longer included). 
Convergence is usually fast, upon which %Upon convergence, 
BP returns an approximation to $\mathbb{E}_{\Phi(Y,X)}~[f_v(X,Y)]$.

As in the E-step, there are some nuances in the M-step when the BP approximation is not exact.
\cite{domke2013learning} contains a more formal discussion of parameter learning with approximate inference methods.
For a more theoretically principled approach, one could use tree-reweighted belief propagation in the M-step and mean-field iteration in the E-step. 
However, we find that our approach is easy to implement and work reasonably well in practice. As aforementioned, the parameter learning problem for the supervision module is much simpler than end-to-end learning with probabilistic logic, as it focuses on refining uncertain weights for task-specific self-supervision, rather than learning complex input patterns for label prediction (handled in deep learning).

\begin{figure}
    \centering
    \includegraphics[width=0.9\linewidth]{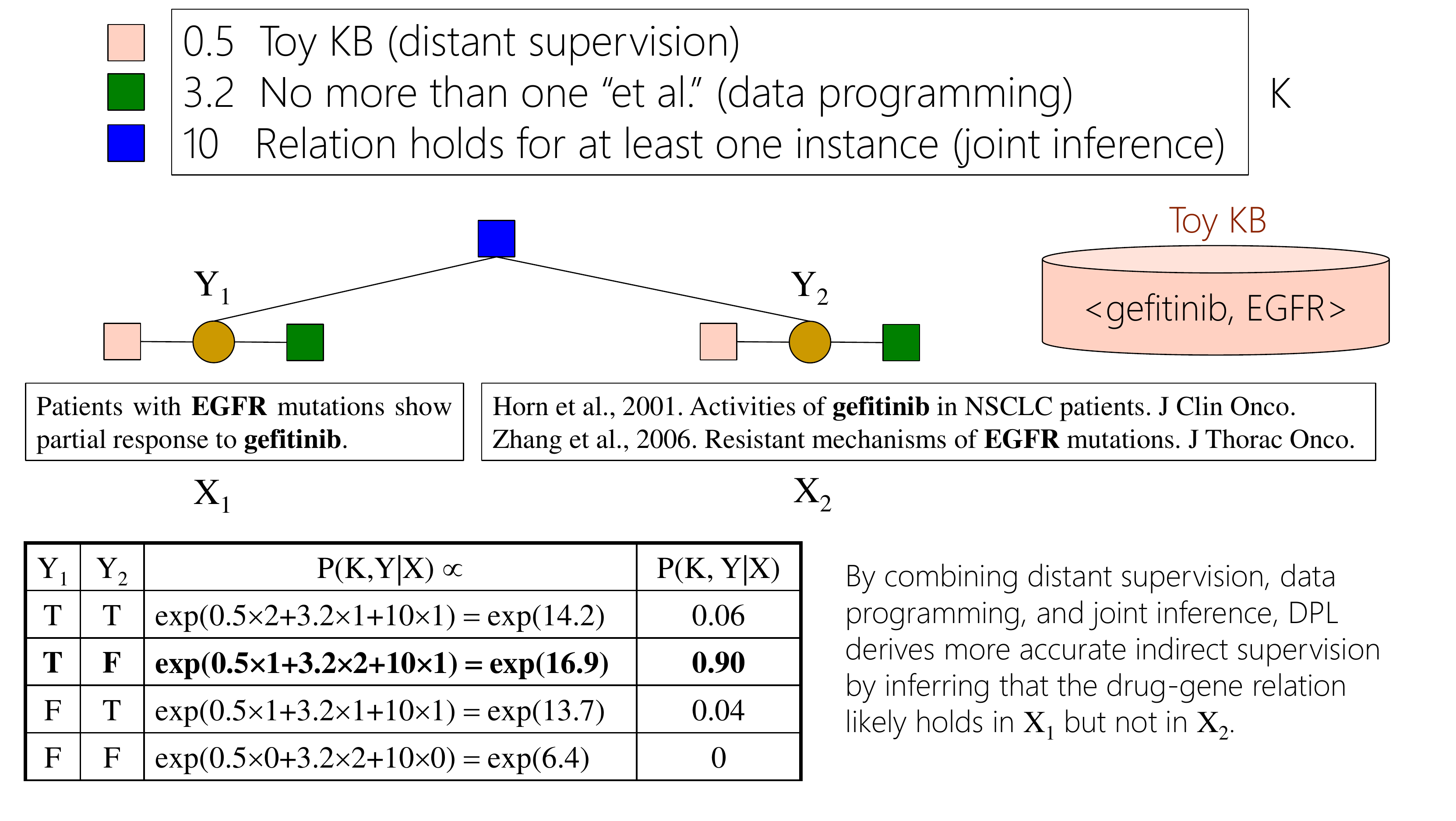}
    \vspace{-10pt}
    \caption{Example of DPL combining various task-specific self-supervision using probabilistic logic. %The prediction module is omitted to avoid clutter. 
    }
    \label{fig:dpl-example}
\end{figure}

\smallpar{Example} 
Figure~\ref{fig:dpl-example} shows a toy example on how DPL combines various task-specific self-supervision for predicting drug-gene interaction (e.g., gefitinib can be used to treat tumors with EGFR mutations). Self-supervision is modeled by probabilistic logic, which defines a joint probability distribution over latent labeling decisions for drug-gene mention pairs in unlabeled text. Here, distant supervision prefers classifying mention pairs of known relations, whereas the data programming formula opposes classifying instances resembling citations, and the joint inference formula ensures that at least one mention pair of a known relation is classified as positive. Formula weight signifies the confidence in the self-supervision, and can be refined iteratively along with the prediction module. 

\smallpar{Handling label imbalance} One challenge for distant supervision is that negative examples are often much more numerous. A common strategy is to subsample negative examples to attain a balanced dataset. In preliminary experiments, we found that this was often suboptimal, as many informative negative examples were excluded from training. Instead, we restored the balance by up-weighting positive examples.
In DPL, an additional challenge is that the labels are probabilistic and change over iterations. In this paper, we simply used hard EM, with binary labels set using 0.5 as the probability threshold, and the up-weighting coefficient recalculated after each E-step. %%

%%%%%%%%%%%%%%%%%%%%%%%%%%%%

\begin{figure}
    \centering
    \includegraphics[width=0.75\linewidth]{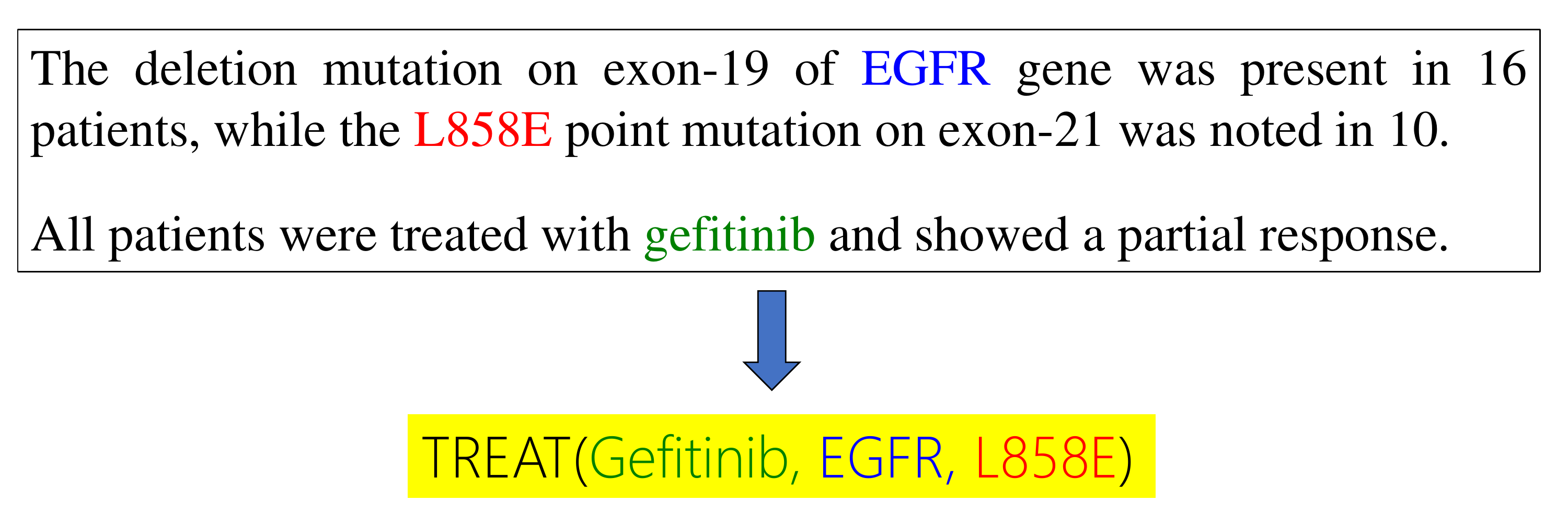} 
    \vspace{-5pt}
    \caption{Example of cross-sentence relation extraction for precision cancer treatment.
    }
    \label{fig:relextract-example}
\end{figure}

\smallsubsection{Biomedical Machine Reading}
\vspace{-5pt}
There is a long-standing interest in biomedical machine reading (e.g., \cite{morgan2008overview, kim2009overview}), but prior studies focused on supervised approaches.
The advent of big biomedical data creates additional urgency for developing scalable approaches that can generalize to new reading tasks. For example, genome sequencing cost has been dropping faster than Moore's Law, yet oncologists can only evaluate tumor sequences for a tiny fraction of patients, due to the bottleneck in assimilating relevant knowledge from publications.
Recently, \cite{peng&al17} formulated precision oncology machine reading as cross-sentence relation extraction (Figure~\ref{fig:relextract-example}) and developed the state-of-the-art system using distant supervision.
While promising, their results still leave much room to improve.
Moreover, they used heuristics to heavily filter entity candidates, with significant recall loss.

In this section, we use cross-sentence relation extraction as a case study for  combining task-specific self-supervision using deep probabilistic logic (DPL).
First, we show that DPL can substantially improve machine reading accuracy in a head-to-head comparison with \cite{peng&al17}, using the same entity linking method. 
Next, we apply DPL to entity linking itself and attain similar improvement.
Finally, we consider further improving the recall by removing the entity filter. 
By applying DPL to joint entity linking and relation extraction, we more than doubled the recall in relation extraction while attaining comparable precision as \cite{peng&al17} with heavy entity filtering.

\smallpar{Evaluation} 
Comparing task-specific self-supervision methods is challenging as there is often
no annotated test set for evaluating precision and recall.
In such cases, we resort to the standard strategy used in prior work by reporting {\em sample precision} (estimated proportion of correct system extractions) and {\em absolute recall} (estimated number of correct system extractions). Absolute recall is proportional to recall and can be used to compare different systems (modulo estimation errors).

\smallpar{Datasets}
We used the same unlabeled text as \cite{peng&al17}, which consists of about one million full text articles in PubMed Central (PMC)\footnote{\url{www.ncbi.nlm.nih.gov/pmc}}.
Tokenization, part-of-speech tagging, and syntactic parsing were conducted using SPLAT \cite{quirk&al12}, and Stanford dependencies \cite{marneffe&al06} were obtained using Stanford CoreNLP \cite{manning&al14}. 
For entity ontologies, we used DrugBank\footnote{\url{www.drugbank.ca}} and Human Gene Ontology (HUGO)\footnote{\url{www.genenames.org}}. 
DrugBank contains 8257 drugs; we used the subset of 599 cancer drugs.
HUGO contains 37661 genes.
For knowledge bases, we used the Gene Drug Knowledge Database (GDKD)~\cite{dienstmann&al15} and the Clinical Interpretations of Variants In Cancer (CIVIC)\footnote{\url{civic.genome.wustl.edu}}. Together, they contain 231 drug-gene-mutation triples, with 76 drugs, 35 genes and 123 mutations.

\begin{table}[t!]
%\footnotesize
\centering
\begin{tabular}[b]{c}
\resizebox{.45\textwidth}{!}{
\begin{tabular}{ |p{8cm} |}
\hline
\textbf{Distant Supervision}: GDKD, CIVIC\\ 
\hline
%\hline
\textbf{Data Programming (Entity)}\\
%\hline
\text{Mention matches entity name exactly.}\\
\text{Mention not a stop word.}\\
\text{Mention not following figure designation.}\\
\text{Mention's POS tags indicate it is a noun.}\\
\hline
%\hline
\textbf{Data Programming (Relation)}\\ 
%\hline
Less than 30\% of words are numbers in each sentence.\\
No more than three consecutive numbers.\\
No more than two ``et al''.\\
No more than three tokens start with uppercase.\\
No more than three special characters.\\
No more than three keywords indicative of table or figure.\\
Entity mentions do not overlap.\\
\hline 
%\hline
\textbf{Joint Inference}: \text{Relation holds in at least one instance.} \\ 
\hline
\end{tabular}
}\\
{\footnotesize (a) Relation Extraction}
\end{tabular}
\hfill
\begin{tabular}[b]{c}
\resizebox{.45\textwidth}{!}{
\begin{tabular}{ | p{8cm} |}
\hline
\textbf{Distant Supervision}: HGNC\\
\hline %\hline
\textbf{Data Programming}\\
No verbs in POS tags.\\
Mention not a common word. \\
Mention contains more than two characters or one word.\\
More than 30\% of characters are upper case. \\
Mention contains both upper and lower case characters. \\
Mention contains both character and digit. \\
Mention contains more than six characters. \\
Dependency label from mention to parent indicative of direct object.\\
\hline %\hline 
\textbf{Joint Inference}\\
Identical mentions nearby probably refer to the same entity.\\
Appositive mentions probably refer to the same entity.\\
Nearby mentions that match synonyms of same entity probably refer to the given entity.
\\ \hline 
\end{tabular}
}\\
{\footnotesize (b) Entity Linking}
\end{tabular}
\caption {DPL combines three types of task-specific self-supervision for relation extraction and entity linking}
\label{tb:factor}	 
\end{table}
%\vspace*{-0.8\baselineskip}

%%%%%%%%%%%%%%%%%%%%%%%%%%%%%%%%%%%%%%%%%%%%%%%%%%%%%%%%%%%%%%%%%%%%%

\smallsubsection{Cross-sentence relation extraction}
\vspace{-5pt}
Let $e_1,\cdots,e_m$ be entity mentions in text $T$. 
Relation extraction can be formulated as classifying whether a relation $R$ holds for $e_1,\cdots,e_m$ in $T$.
To enable a head-to-head comparison, we used the same cross-sentence setting as \cite{peng&al17}, where $T$ spans up to three consecutive sentences and $R$ represents the ternary interaction over drugs, genes, and mutations (whether the drug is relevant for treating tumors with the given gene mutation).

\smallpar{Entity linking}
In this subsection, we used the entity linker from Literome \cite{poon&al14} to identify drug, gene, and mutation mentions, as in \cite{peng&al17}. 
This entity linker first identifies candidate mentions by matching entity names or synonyms in domain ontologies, then applies heuristics to filter candidates. The heuristics are designed to enhance precision, at the expense of recall. For example, one heuristics would filter candidates of length less than four, which eliminates key cancer genes such as ER or AKT.

\smallpar{Prediction module}
We used the same graph LSTM as in \cite{peng&al17} to enable head-to-head comparison on self-supervision.
Briefly, a graph LSTM generalizes a linear-chain LSTM by incorporating arbitrary long-ranged dependencies, such as syntactic dependencies, discourse relations, coreference, and connections between roots of adjacent sentences.
A word might have precedents other than the prior word, and its LSTM unit is expanded to include a forget gate for each precedent. 
See \cite{peng&al17} for details.

\smallpar{Supervision module}
We used DPL to combine three types of task-specific self-supervision for cross-sentence relation extraction (Table~\ref{tb:factor} (a)).
For distant supervision, we used GDKD and CIVIC as in \cite{peng&al17}.
For data programming, we introduced labeling functions that aim to correct entity and relation errors.
Finally, we incorporated joint inference among all co-occurring instances of an entity tuple with the known relation by imposing the at-least-one constraint (i.e., the relation holds for at least one of the instances).
For development, we sampled 250 positive extractions from DPL using only distant supervision \cite{peng&al17} and excluded them from future training and evaluation.

\begin{table}[t]
\centering
\begin{tabular}[b]{c}
\resizebox{0.45\textwidth}{!}{
\begin{tabular}{ | l | c | c | c |}
\hline
System & Prec. & Abs. Rec.  & Unique \\ \hline
Peng 2017  & 0.64 & 6768 & 2738 \\ \hline
DPL + $\tt EMB$ & {\bf 0.74} & {\bf 8478} &  {\bf 4821} \\ \hline
DPL & 0.73 & 7666 &  4144 \\ \hline
~~~$-$ $\tt DS$ & 0.29 & 7555 & 4912  \\ \hline
~~~$-$ $\tt DP$ & 0.67 & 4826 &  2629 \\ \hline
~~~$-$ $\tt DP$ $\tt (ENTITY)$ & 0.70 & 7638 &  4074 \\ \hline
~~~$-$ $\tt JI$ & 0.72 & 7418 & 4011  \\ \hline
\end{tabular}
}\\
{\footnotesize (a)}
\end{tabular}
\begin{tabular}[b]{c}
\resizebox{0.45\textwidth}{!}{
\begin{tabular}{ | l | c | c | c |}
\hline
Pred. Mod. & Prec. & \ Abs. Rec. & Unique \\ \hline
BiLSTM  & 0.60 & 6243 &  3427 \\ \hline
Graph LSTM  & 0.73 & 7666 &  4144 \\ \hline
\end{tabular}
}\\
{\footnotesize (b)}
\end{tabular}
\caption {
Comparison of sample precision and absolute recall in test extraction on PMC. (a) Ablation study on self-supervision (DS: distant supervision, DP: data programming, JI: joint inference). 
DPL + $\tt EMB$ is our system using PubMed-trained word embedding, whereas DPL uses original Wikipedia-trained word embedding in \cite{peng&al17}. (b) Comparison of prediction model, with same set of self-supervision and Wikipedia-trained word embedding.
}
\label{tbl:RE-ablation}
\end{table}
%\vspace*{-0.8\baselineskip}

\smallpar{Experiment results}
We compared DPL with the state-of-the-art system of \cite{peng&al17}. We also conducted ablation study to evaluate the impact of self-supervision strategies.
For a fair comparison, we used the same probability threshold in all cases (an instance is classified as positive if normalized probability score is at least 0.5).
For each system, sample precision was estimated by sampling 100 positive extractions and manually determining the proportion of correct extractions by an author knowledgeable about this domain.
Absolute recall is estimated by multiplying sample precision with the number of positive extractions. 
Table~\ref{tbl:RE-ablation} (a) shows the results. DPL substantially outperformed \cite{peng&al17}, improving sample precision by ten absolute points and raising absolute recall by 25\%.
Combining self-supervision is key to this performance gain, as evident from the ablation results. While distant supervision remained the most potent source of task-specific self-supervision, data programming and joint inference each contributed significantly.
Replacing out-of-domain (Wikipedia) word embedding with in-domain (PubMed) word embedding~\cite{pyysalo&al13} also led to a small gain. 
\cite{peng&al17} only compared graph LSTM and linear-chain LSTM in automatic evaluation, where distant-supervision labels were treated as ground truth. They found significant but relatively small gains by graph LSTM. We conducted additional manual evaluation comparing the two in DPL. 
Surprisingly, we found large performance difference, with graph LSTM outperforming linear-chain LSTM by 13 absolute points in precision and raising absolute recall by over 20\% (Table~~\ref{tbl:RE-ablation} (b)). 
This suggests that \cite{peng&al17} might have underestimated performance gain using automatic evaluation.

%%%%%%%%%%%%%%%%%%%%%%%%%%%%%%%%%%%%%%%%%%%%%%%%%%%%%%%%%%%%%%%%%%%%%%

\smallsubsection{Entity linking}
\vspace{-5pt}
Let $m$ be a mention in text and $e$ be an entity in an ontology. The goal of entity linking is to predict $\tt Link(m,e)$, which is true iff $m$ refers to $e$, for every candidate mention-entity pair $m,e$.
We focus on genes in this paper, as they are particularly noisy.

\smallpar{Prediction module} We used BiLSTM with attention over the ten-word windows before and after a mention.
The embedding layer is initialized by word2vec embedding trained on PubMed text~\cite{pyysalo&al13}.
The word embedding dimension was 200. We used 5 epochs for training, with Adam as the optimizer. We set learning rate to 0.001, and batch size to 64. 

\smallpar{Supervision module}
As in relation extraction, we combined three types of self-supervision using DPL (Table~\ref{tb:factor} (b)).
For distant supervision, we obtained all mention-gene candidates by matching PMC text against the HUGO lexicon. We then sampled a subset of 200,000 candidate instances as positive examples. We sampled a similar number of noun phrases as negative examples. 
For data programming, we introduced labeling functions that used mention characteristics (longer names are less ambiguous) or syntactic context (genes are more likely to be direct objects and nouns). 
For joint inference, we leverage linguistic phenomena related to coreference (identical, appositive, or synonymous mentions nearby are likely coreferent).

\begin{table}[t]
\centering
\begin{tabular}[b]{c}
	\resizebox{0.45\textwidth}{!}{
		\begin{tabular}{ | l | c | c | c | c| }
			\hline
			System & Acc. & F1 & Prec. & Rec. \\ \hline
			String Match & 0.18 & 0.31 & 0.18 & 1.00 \\ \hline
			DS & 0.64 & 0.71 & 0.62 & 0.83 \\ \hline
			DS + DP   & 0.66 & 0.71 & 0.62 & 0.83\\ \hline
			DS + DP + JI & {\bf 0.70} & {\bf 0.76}   & 0.68 & 0.86   \\ \hline
		\end{tabular}
		}\\
		{\footnotesize (a)}
		\end{tabular}
\begin{tabular}[b]{c}
    \resizebox{0.45\textwidth}{!}{
    \begin{tabular}{ | l | c | c | c |}
    \hline
     & F1 & Precision & Recall \\ \hline
    GNormPlus  & 0.78 & 0.74 & 0.81 \\ \hline
    DPL  & 0.74 & 0.68 & 0.80 \\ \hline
    \end{tabular}
    }\\
	{\footnotesize (b)}
	\end{tabular}
\caption {(a) Comparison of gene entity linking results on a balanced test set. The string-matching baseline has low precision. By combining various types of task-specific self-supervision, DPL substantially improved precision while retaining reasonably high recall. (b) Comparison of gene entity linking results on BioCreative II test set. GNormPlus is the state-of-the-art system trained on thousands of labeled examples. DPL used only task-specific self-supervision.
}
\label{tab:EL}	 
\end{table}

\smallpar{Experiment results}
For evaluation, we annotated a larger set of sample gene-mention candidates and then subsampled a balanced test set of 550 instances (half are true gene mentions, half not).
These instances were excluded from training and development.
Table~\ref{tab:EL} (a) compares system performance on this test set.
The string-matching baseline has a very low precision, as gene mentions are highly ambiguous, which explains why \cite{peng&al17} resorted to heavy filtering.
By combining various task-specific self-supervision, DPL improved precision by over 50 absolute points, while retaining a reasonably high recall (86\%). All self-supervision contributed significantly, as the ablation tests show.
We also evaluated DPL on BioCreative II, a shared task on gene entity linking  \cite{morgan2008overview}. 
We compared DPL with GNormPlus \cite{wei2015gnormplus}, the state-of-the-art supervised system trained on thousands of labeled examples in BioCreative II training set.
Despite using zero manually labeled examples, DPL attained comparable F1 and recall (Table~\ref{tab:EL} (b)). The difference is mainly in precision, which indicates opportunities for more task-specific self-supervision.

\smallsubsection{Joint entity and relation extraction}
\vspace{-5pt}
An important use case for machine reading is to improve knowledge curation efficiency by offering extraction results as candidates for curators to vet. The key to practical adoption is attaining high recall with reasonable precision~\cite{peng&al17}.
The entity filter used in \cite{peng&al17} is not ideal in this aspect, as it substantially reduced recall. 
In this subsection, we consider replacing the entity filter by the DPL entity linker. See Table~\ref{tab:joint} (a). 
Specifically, we added one labeling function to check if the entity linker returns a normalized probability score above $p_{\tt TRN}$ for gene mentions, and filtered test instances if the gene mention score is lower than $p_{\tt TST}$. 
We set $p_{\tt TRN}=0.6$ and $p_{\tt TST}=0.3$ from preliminary experiments. The labeling function discouraged learning from noisy mentions, and the test-time filter skips an instance if the gene is likely wrong. 
Not surprisingly, without entity filtering, \cite{peng&al17} suffered large precision loss.
All DPL versions substantially improved accuracy, with significantly more gains using the DPL entity linker.

\begin{table}[t]
\centering
\begin{tabular}[b]{c}
\resizebox{0.45\textwidth}{!}{
\begin{tabular}{ | l | c | c | c |}
\hline
System & Prec & Abs. Rec. & Unique \\ \hline
Peng 2017 &	0.31 & 11481 & 5447 \\ \hline
DPL (RE)  & 0.52 & 17891 & 8534 \\ \hline
~$+$ EL (TRN) & 0.55 & {\bf 21881} &  {\bf 11047} \\ \hline
~$+$ EL (TRN/TST) & {\bf 0.61} & 20378 &  10291 \\ \hline
\end{tabular}
}\\
{\footnotesize (a)}
\end{tabular}
\begin{tabular}[b]{c}
\resizebox{0.45\textwidth}{!}{
\begin{tabular}{ | c | c | c| c | c|}
\hline
Gene & Drug & Mut. & Gene-Mut. & Relation\\
\hline
27\% & 4\% & 20\% & 45\% & 24\% \\ \hline
\end{tabular}
}\\
{\footnotesize (b)}
\end{tabular}
\caption{(a) Comparison of sample precision and absolute recall (all instances and unique entity tuples) when all gene mention candidates are considered. \cite{peng&al17} used distant supervision only. RE: DPL relation extraction. EL: using DPL entity linking in RE training (TRN) and/or test (TST). (b) Error analysis for DPL relation extraction.
}
\label{tab:joint}	 
\end{table}

\begin{figure}
    \centering
    \includegraphics[width=0.7\linewidth]{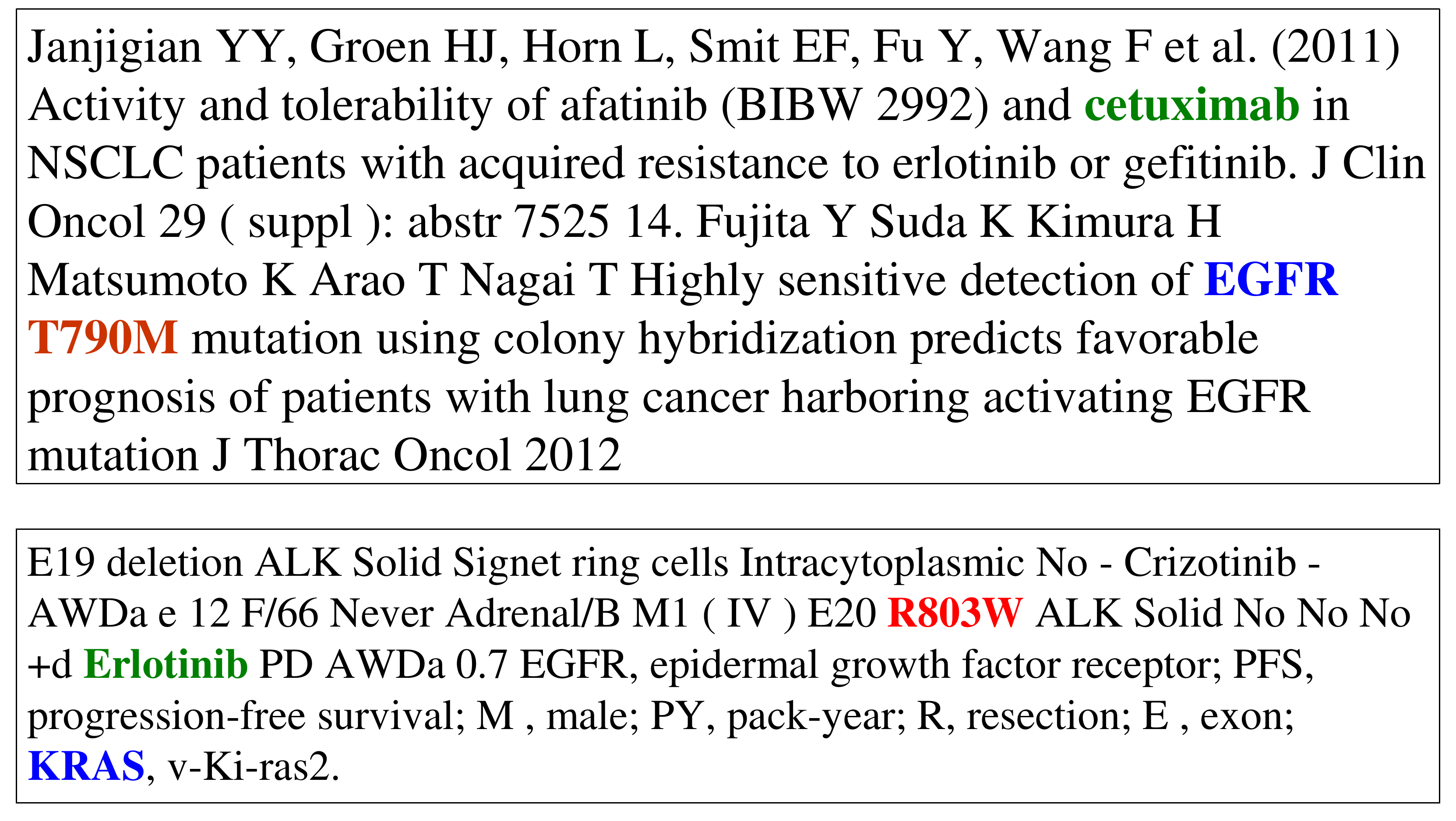}
    \vspace{-10pt}
    \caption{Example of relation-extraction errors corrected by DPL with additional task-specific self-supervision. 
    }
    \label{fig:fix-example}
\end{figure}

\smallsubsection{Discussion}
\vspace{-5pt}

\smallpar{Scalability} 
DPL is efficient to train, taking around 3.5 hours for relation extraction and 2.5 hours for entity linking in our PubMed-scale experiments, with 25 CPU cores (for probabilistic logic) and one GPU (for LSTM). For relation extraction, the graphical model of probabilistic logic contains around 7,000 variables and 70,000 factors. At test time, it is just an LSTM, which predicted each instance in less than a second. 
In general, DPL learning scales linearly in the number of training instances. For distant supervision and data programming, DPL scales linearly in the number of known facts and labeling functions. As discussed before, joint inference with high-order factors is more challenging, but can be efficiently approximated. For inference in probabilistic logic, we found that loopy belief propagation worked reasonably well, converging after 2-4 iterations. Overall, we ran variational EM for three iterations, using ten epochs of deep learning in each M-step. We found these worked well in preliminary experiments and used the same setting in all final experiments.

\smallpar{Accuracy}
To understand more about DPL's performance gain over distant supervision, we manually inspected some relation-extraction errors fixed by DPL after training with additional self-supervision. 
Figure~\ref{fig:fix-example} shows two such examples. 
While some data programming functions were introduced to prevent errors stemming from citations or flattened tables, none were directly applicable to these examples. This shows that DPL can generalize beyond the original self-supervision.

While the results are promising, there is still much to improve. 
Table~\ref{tab:joint} (b) shows estimated precision errors for relation extraction by DPL. (Some instances have multiple errors.) 
Entity linking can incorporate more task-specific self-supervision.
Joint entity linking and relation extraction can be improved by feeding back extraction results to linking. 
Improvement is also sorely needed in classifying mutations and gene-mutation associations.
The prediction module can also be improved, e.g., by adding attention to graph LSTM.
DPL offers a flexible framework for exploring all these directions.

\smallsection{Self-Supervised Self-Supervision}
\label{sec:s4}
\vspace{-5pt}

\begin{algorithm}[t]
\footnotesize
\begin{algorithmic}
\caption{Self-Supervised Self-Supervision (S4)}\label{alg:S4}
\State \textbf{Input:} Seed virtual evidences $I$, deep neural network $\Psi$, inputs $X=(X_1,\ldots,X_N)$, unobserved outputs $Y=(Y_1,\ldots,Y_N)$, human query budget $T$.
\State \textbf{Output:} Learned prediction module $\Psi$ and virtual evidences $K = \{(f_v(X,Y), w_v):v\}$.
\State \textbf{Initialize:} $K=I$;  $Q=\emptyset$; $i  = 0$.
%\For{$i = 1..M$; $i<M$; $i$\small{++}}
\For{$i = 1\ldots M$}{
\While{Structured Self-Training not converged}{
    \State \qquad\qquad $\Psi, K \gets \mbox{\texttt{DPL-Learn}}(K, \Psi, X, Y)$
    \State \qquad\qquad $v=\texttt{PropSST}(K, \Psi, X, Y)$; 
    \State \qquad\qquad $K\gets K\cup v$; 
}
\If{$|Q|<T$}{
\State \qquad\qquad $v=\mbox{\texttt{PropFAL}}(K, \Psi, X, Y, Q); Q\gets Q \cup v$;
\State \qquad\qquad  {\bf if} $\mbox{\texttt{Human-Accept}}(v)$ {\bf then} $K \gets K \cup v$
}
}
\end{algorithmic}
\end{algorithm}

%% overview
In this section, we present the {\em Self-Supervised Self-Supervision (S4)} framework, which extends deep probabilistic logic (DPL) with the capability to learn new self-supervision.
Let~$\mathcal{V}=\{(f_v,w_v,\alpha_v):v\}$~be the set of all candidate virtual evidences, where $f_v(X,Y)$ is a first-order logical formula, $w_v$ is the weight, and $\alpha_v$ is the weight prior (for non-fixed $w_v$).
Let $K$ be the set of virtual evidences maintained by the algorithm, initialized by the seed $I$. 
The key idea of S4 is to interleave structure learning and active learning to iteratively propose new virtual evidence $v\in \mathcal{V}$ to augment $K$ (Figure~\ref{fig:S4}). 

Specifically, S4 can be viewed as conducting {\em structure learning} in the factor graph that specifies the virtual evidence. Structure learning has been studied intensively in the graphical model literature \cite{koller-struc-lrn}. It is also known as feature selection or feature induction in general machine learning literature \cite{feat-sel}.
Here, we are introducing structured factors for self-supervision, rather than as feature templates to be used during training. 
Another key difference from standard structure learning is the deep neural network, which provides an alternative view from the virtual evidence space and enables multi-view learning in DPL. The neural network can also help identify candidate virtual evidences, e.g., via neural attention. 
Self-training is a special case where candidate virtual evidences are individual label assignments (i.e., $f_v=I[y_v=l_v]$).
S4 can thus be viewed as conducting {\em structured self-training (SST)} by admitting arbitrary Markov logic formulas as virtual evidence.

In data programming and other prior self-supervision methods, human experts need to pre-specify all self-supervision upfront. While it is easy to generate a small seed by identifying the most salient self-supervision, this effort can quickly become tedious and more challenging as the experts are required to enumerate the less salient templates. On the other hand, given a candidate, it's generally much easier for experts to validate it. {\em This suggests that for the best utilization of human bandwidth, we should focus on leveraging them to produce the initial self-supervision and verify candidate self-supervision.}
Consequently, in addition to structured self-training (SST), S4 incorporates {\em feature-based active learning (FAL)} (i.e., active learning of self-supervision). When SST converges, S4 will switch to the active learning mode by proposing a candidate virtual evidence for human verification (i.e., labeling a feature rather than an instance in standard active learning). Intuitively, in FAL we are proposing virtual evidences for which the labels of the corresponding instances are still uncertain. If the human expert can provide definitive supervision on the label, the information gain will be large.
By contrast, in SST, we favor virtual evidences with skewed posterior label assignments for their corresponding instances, as they can potentially amplify the signal.

Algorithm~\ref{alg:S4} describes the S4 algorithm. S4 first conducts DPL using the initial self-supervision $I$, then interleaves structured self-training (SST) with feature-based active learning (FAL). 
SST steps are repeated until there is little change in the probabilistic labels ($<1\%$ in our experiments).
DPL learning updates the deep neural network and the graphical model parameters with warm start (i.e., the parameters are initialized with the previous parameters). 
All proposed queries are stored and won't be proposed again.
The total amount of human effort consists of generating the seed $I$ and validating $T$ queries. 

S4 is a general algorithmic framework that can combine various strategies for designing $\mathcal{V}$, $\tt PropSST$, and $\tt PropFAL$.
In standard structure learning, $\tt PropSST$ would attempt to maximize the learning objective (e.g., conditional likelihood of seed virtual evidences) by iteratively conducting greedy structure changes. However, this is very expensive to compute, since it requires a full DPL run just to score each candidate.
Instead, we take inspiration from the feature-induction and relational learning literature and use heuristic approximations that are much faster to evaluate.
%% VE classes
In the most general setting, $\cV$ contains all possible potential functions. In practice, we can restrict it to obtain a good trade-off between expressiveness and computation for the problem domain. 
Interestingly, as we will see in the experiment section, even with relatively simple classes of self-supervision, S4 can dramatically improve over DPL through structure learning and active learning. 
We use text classification from natural language processing (NLP) as a running example to illustrate how to apply S4 in practice. 
Here, the input $X_i=(t_1,\ldots,t_{s_i})$ is a sequence of tokens and the output $Y_i$ is the classification label (e.g., $\tt pos$ or $\tt neg$ in sentiment analysis).

\smallsubsection{Candidate Self-Supervision}
\vspace{-5pt}
For $\mathcal{V}$, the simplest choice is to use tokens. Namely, $f_{t,l}(X_i,Y_i)=\mathbb{I}[t\in X_i\land Y_i=l]$. For simplicity, we can use a fixed weight and prior for all initial virtual evidence, i.e., $\cV=\{(f_{t,l},w,\alpha): t,l\}$. 
Take sentiment analysis as an example. $X_i$ may represent a movie review and $Y_i\in\{0,1\}$ the sentiment. 
A virtual evidence for self-supervision may stipulate that if the review contains the word ``good'', the sentiment is more likely to be positive. This can be represented by the formula $f_{\text{good},1}(X_i,Y_i) = \mathbb{I}[\text{``good''} \in X_i \wedge Y_i=1]$ with a positive weight.
A more advanced choice for $\cV$ may include high-order joint-inference factors, such as $f_{ij}(Y_i,Y_j)=\mathbb{I}[Y_i=Y_j]$. If we add this factor for similar pairs $X_i, X_j$, it stipulates that instances with similar input are likely to share the same label.
Here we define similar pairs with a function ${\tt Sim}(X_i,X_j)$ between $X_i$ and $X_j$, such as the cosine similarity between the sentence (or document) embeddings of $X_i$ and $X_j$, based on the current deep neural network.
Note that this is different from graph-based semi-supervised learning or other kernel-based methods in that the similarity metric is not pre-specified and fixed, but rather evolving along with the neural network for the end task.

\smallsubsection{Structured Self-Training ($\tt PropSST$)} 
\vspace{-5pt}
From DPL learning, we obtain the current marginal estimate of the latent label variables $q_i(Y_i)$, which we would treat as probabilistic labels in assessing candidate virtual evidence. 
There are many sensible strategies for proposing candidates in structured self-training (i.e., $\tt PropSST$).
For token-based self-supervision, a common technique from the feature-selection literature is to choose a token highly correlated with a label.
For example, we can choose the token $t$ that occurs much more frequently in instances for a given label $l$ than others using our noisy label estimates.
We find that this often leads to very noisy proposals and semantic drift.
A simple refinement is to restrict our scoring to instances containing some initial self-supervised tokens.
However, this still has the drawback that a word may occur more often in instances of a class for reasons other than contributing to the label classification.
We therefore consider a more sophisticated strategy based on neural attention.
Namely, we will credit occurrences using the normalized attention weight for the given token in each instance.

Formally, let $A_{\Psi}(X_i,j)$ represent the normalized attention weight the neural network $\Psi$ assigns to the $j$-th token in $X_i$ for the final classification. We define {\em average weighted attention} for token $t$ and label $l$ as ${\tt Attn}(t,l)=\frac{1}{C_t}\sum_{i,j: X_{i,j}=t}~q_i(Y_i=l)\cdot A_{\Psi}(X_i,j)$, where $C_t$ is the number of occurrences of $t$ in $X$. 
Then $\tt PropSST$ would simply score token-based self-supervision $f_{t,l}$ using relative average weighted attention: $S_{\text{token}}(t,l)={\tt Attn}(t,l)-\sum_{l'\ne l}{\tt Attn}(t,l')$. 
In each iteration, $\tt PropSST$ picks the top scoring $f_{t,l}$ that has not been proposed yet as the new virtual evidence to add to $K$.

We also consider an entropy-based score function that works for arbitrary input-based features. It treats the prediction module $\Psi$ as a black box, and only uses the posterior label assignments $q_i(Y_i)$. 
Consider candidate virtual evidence $f_{b,l}(X_i,Y_i) = \mathbb{I}[b(X_i) \wedge Y_i = l]$, where $b$ is a binary function over input $X_i$. This clearly generalizes token-based virtual evidence. 
Define ${\tt Ent}(b) = H\left(\frac{1}{C_b}\sum_{i: b(X_i) = 1} q_i(Y_i)\right)$,  where $H$ is the Shannon entropy and $C_b$ is the number of instances for which the feature $b$ holds true. This function represents the entropy of the average posterior among all instances with $b(X_i)=1$.
$\tt PropSST$ will then use $S_\text{entropy}(b)=1/{\tt Ent}(b)$ to choose the $b^*$ with the lowest average entropy and then pick label $l^*$ with the highest average posterior probability for $b^*$. In our experiments, this performs similarly to attention-based scores.

For joint-inference self-supervision, we consider the similarity-based factors defined earlier, and leave the exploration of more complex factors to future work.
To distinguish task-specific similarity from pretrained similarity, we use the difference between the similarity computed using the current fine-tuned BERT model and that using the pretrained one.
Formally, let ${\tt Sim}_{\text{pretrained}}(X_i,X_j)$ be the cosine similarity between the embeddings of $X_i$ and $X_j$ generated by the pretrained BERT model, and ${\tt Sim}_{\Psi}(X_i,X_j)$ be that between the embeddings using the current learned network $\Psi$. $\tt PropSST$ would score the joint-inference factor using the relative similarity and choose the top scoring pairs to add to self-supervision:
$S_{\text{joint}}(X_i,X_j) = {\tt Sim}_{\Psi}(X_i,X_j) - {\tt Sim}_{\text{pretrained}}(X_i,X_j)$.

\smallsubsection{Feature-Based Active Learning ($\tt PropFAL$)} 
\vspace{-5pt}
For active learning, a common strategy is to pick the instance with highest entropy in the label distribution based on the current marginal estimate. In feature-based active learning, we can similarly pick the feature $b$ with the highest average entropy ${\tt Ent}(b)$. Note that this is opposite to how we use the entropy-based score function in $\tt PropSST$, where we choose the feature with the lowest average entropy.
In $\tt PropFAL$, we will identify $b^*=\argmax ({\tt Ent}(b))$, present $f_{b^*,l}(X,Y)=\mathbb{I}[b^*(X)\land Y=l]$ for all possible labels $l$, and ask the human expert to choose a label $l^*$ to accept or reject them all. 

%%%%%%%%%%%%%%%%%%%%%%%%%%%%%%%%%%%%%%%%%%%%%%%%%%%%%%

\smallsubsection{Experiments}
\label{sec:experiments}
\vspace{-5pt}
We use the natural language processing (NLP) task of text classification to explore the potential for S4 to improve over DPL using structure learning and active learning. We used three standard text classification datasets: IMDb \cite{maas2011learning}, Stanford Sentiment Treebank \cite{socher2013recursive}, and Yahoo!~Answers \cite{zhang2015character}.
IMDb contains movie reviews with polarity labels (positive/negative). There are 25,000 training instances with equal numbers of positive and negative labels, and the same numbers for test. 
Stanford Sentiment Treebank (StanSent) also contains movie reviews, but was annotated with five labels ranging from very negative to very positive. We used the binarized version of StanSent, which collapses the polarized categories and discards the neutral sentences. It contains 6,920 training instances and 1,821 test instances, with roughly equal split.
Overall, the StanSent reviews are shorter than IMDb's, and they often exhibit more challenging linguistic phenomena (e.g., nested negations or sarcasm). The Yahoo dataset contains 1.4 million training questions and 60,000 test questions from Yahoo!~Answers; these are equally split into 10 classes. %Due to space constraints, we report the Yahoo results in the appendix.

%% implementation details
In all our experiments with S4, we withheld gold labels from the system, used the training instances as unlabeled data, and evaluated on the test set. We reported test accuracy, as all of the datasets are class-balanced.
For our neural network prediction module $\Psi(X_i,Y_i)$, we used the standard BERT-base model pretrained using Wikipedia \cite{devlin2018bert}, along with a global-context attention layer as in \cite{yang2016hierarchical}, which we also used for attention-based scoring. 
We truncated the input text to 512 tokens, the maximum allowed by the standard BERT model.
All of our baselines (except supervised bag-of-words) use the same BERT model.
For all virtual evidences, we used initial weight $w=2.2$ (the log-odds of 90\% probability) and used an $\alpha$ corresponding to an L2 penalty of $5\times10^{-8}$ on $w$. Our results are not sensitive to these values.
In all experiments, we use the Adam optimizer with an initial learning rate tuned over $[0.1, 0.01, 0.001]$. The optimizer's history is reset after each EM iteration to remove old gradient information. We always performed 3 EM iterations and trained $\Psi$ for 5 epochs per iteration. 
For the global-context attention layer, we used a context dimension of 5. The model is warm-started across EM iterations (in DPL), but \emph{not} across the outer iterations in S4 (the for loop). In all experiments, we used the Adam optimizer with an initial learning rate tuned over $[0.1, 0.01, 0.001]$. The optimizer's history is reset after each EM iteration to remove old gradient information. 

%% VE & Simulate human supervision
For virtual evidence, we focus on token-based unary factors and similarity-based joint factors, as discussed in the previous section, and leave the exploration of more complex factors to future work. Even with these factors, our self-supervised $\Psi$ models often nearly match the accuracy of the best supervised models. We also compare with Snorkel, a popular self-supervision system \cite{ratner2016data}. We use the latest Snorkel version  \cite{ratner2019training}, which models correlations among same-instance factors. Snorkel cannot incorporate joint-inference factors across different instances. In all of our Snorkel baselines, we separately tuned the initial learning rate over the same set, and trained the deep neural network for the same number of {\em total} epochs that DPL uses to ensure a fair comparison.

To simulate human supervision for unary factors, we trained a unigram model using the training data with L1 regularization and selected the 100 tokens with the highest weights for each class as the oracle self-supervision. 
By default, we used the top tokens for each class in the initial self-supervision $I$. We also experimented with using random tokens from the oracle in $I$ to simulate lower-quality initial supervision and to quantify the variance of S4.
For the set of oracle joint factors, we fine-tuned the standard BERT model on the training set, used the $\tt CLS$ embedding BERT produces to compute input similarity, and picked the 100 input pairs whose similarity changed the most between the fine-tuned model and the initial model. 

\begin{table}
\centering
\begin{tabular}[b]{c}
      \resizebox{0.45\textwidth}{!}{
        \begin{tabular}{lcc}
            \toprule
            Algorithm   & Sup. size $|I|$  & Test acc (\%) \\
            \midrule
            BoW & 25k &  87.1 \\
            DNN & 25k & 91.0  \\
            \midrule
            \multirow{2}{*}{Self-training} & 100 &  69.9   \\
             & 1k &  88.5   \\
            \midrule            
            Snorkel & 6 & 76.6\\
            DPL & 6 &  80.7   \\
            S4-SST & 6     &  85.5 \\
            S4 ($T=20$) & 6 & 85.6\\
            \midrule            
            Snorkel & 20 & 82.4 \\
            DPL & 20           & 78.9  \\
            S4-SST & 20     &  86.4 \\
            S4 ($T=20$) & 20 & 86.9\\
            \midrule           
            Snorkel & 40 & 84.6\\
            DPL & 40 & 85.2\\
            S4-SST & 40 & 86.6 \\
            S4 ($T=20$) & 40 & 86.8\\
            \bottomrule
        \end{tabular}
        } \\
        {\footnotesize (a) IMDb}
\end{tabular}
\begin{tabular}[b]{c}
      \resizebox{0.45\textwidth}{!}{
        \begin{tabular}{lcc}
            \toprule
            Algorithm & Sup. size $|I|$    & Test acc (\%) \\
            \midrule
            BoW & 6.9k & 78.9  \\
            DNN & 6.9k & 90.9 \\
            \midrule
            \multirow{2}{*}{Self-training}   & 50  & 78.0  \\
              & 100  & 81.8  \\
            \midrule
            Snorkel & 6 & 63.5\\            
            DPL & 6 & 57.2\\
            S4-SST  & 6 & 73.0\\
            S4-SST + J & 6 & 76.2\\
            S4 + J ($T=20$) & 6 & 81.4\\
            \midrule
            Snorkel & 20 & 73.0\\
            DPL & 20 & 72.4\\            
            S4-SST & 20 & 83.3\\
            S4-SST + J & 20 & 85.1\\
            S4 + J ($T=20$) & 20 & 84.4 \\
            \midrule
            Snorkel & 40 & 73.6\\            
            DPL & 40 & 77.0\\
            S4-SST & 40 & 84.9\\
            S4-SST + J & 40 & 86.3\\
            S4 + J $(T=20)$ & 40 & 85.4\\
            \bottomrule
        \end{tabular}
        } \\
        {\footnotesize (b) Stanford}
\end{tabular}
    \caption{System comparison on IMDb and Stanford.}\label{tbl:sys}      
\end{table}

%% structure learning
We first investigate whether structure learning can help in S4 by running without feature-based active learning. We set the query budget $T=0$ in Algorithm \ref{alg:S4}. Because we only take structured self-training steps when $T=0$, we denote this version of S4 as S4-SST.
Table~\ref{tbl:sys} (a) shows the results on IMDb.
With just six self-supervised tokens (three per class), S4-SST already attained 86\% test accuracy, which outperforms self-training with 100 labeled examples by 16 absolute points, and is only slightly worse than self-training with 1000 labeled examples or supervised training with 25,000 labeled examples.
By conducting structure learning, S4-SST substantially outperformed DPL, gaining about 5 absolute points in accuracy (a 25\% relative reduction in error), and also outperformed the Snorkel baseline by 8.9 points. 
Interestingly, with more self-supervision at 20 tokens, DPL's performance drops slightly, which might stem from more noise in the initial self-supervision. By contrast, S4-SST capitalized on the larger seed self-supervision and attained steady improvement.
Even with substantially more self-supervision at 40 tokens, S4-SST still attained similar accuracy gain, demonstrating the power in structure learning.
On average across different initial amounts of supervision $|I|$, S4-SST outperforms DPL by 5.6 points and Snorkel by 5.3 points. 
Next, we consider the full S4 algorithm with a budget of up to $T=20$ human queries (S4 ($T=20$)). Overall, by automatically generating self-supervision from structure learning, S4-SST already attained very high accuracy on this dataset. However, active learning can still produce some additional gain. 
The only randomness in Table \ref{tbl:sys} is the initialization of the deep network $\Psi$, with negligible effect.

\begin{figure}
\centering
\begin{tabular}[b]{c}
    \includegraphics[width=0.45\textwidth]{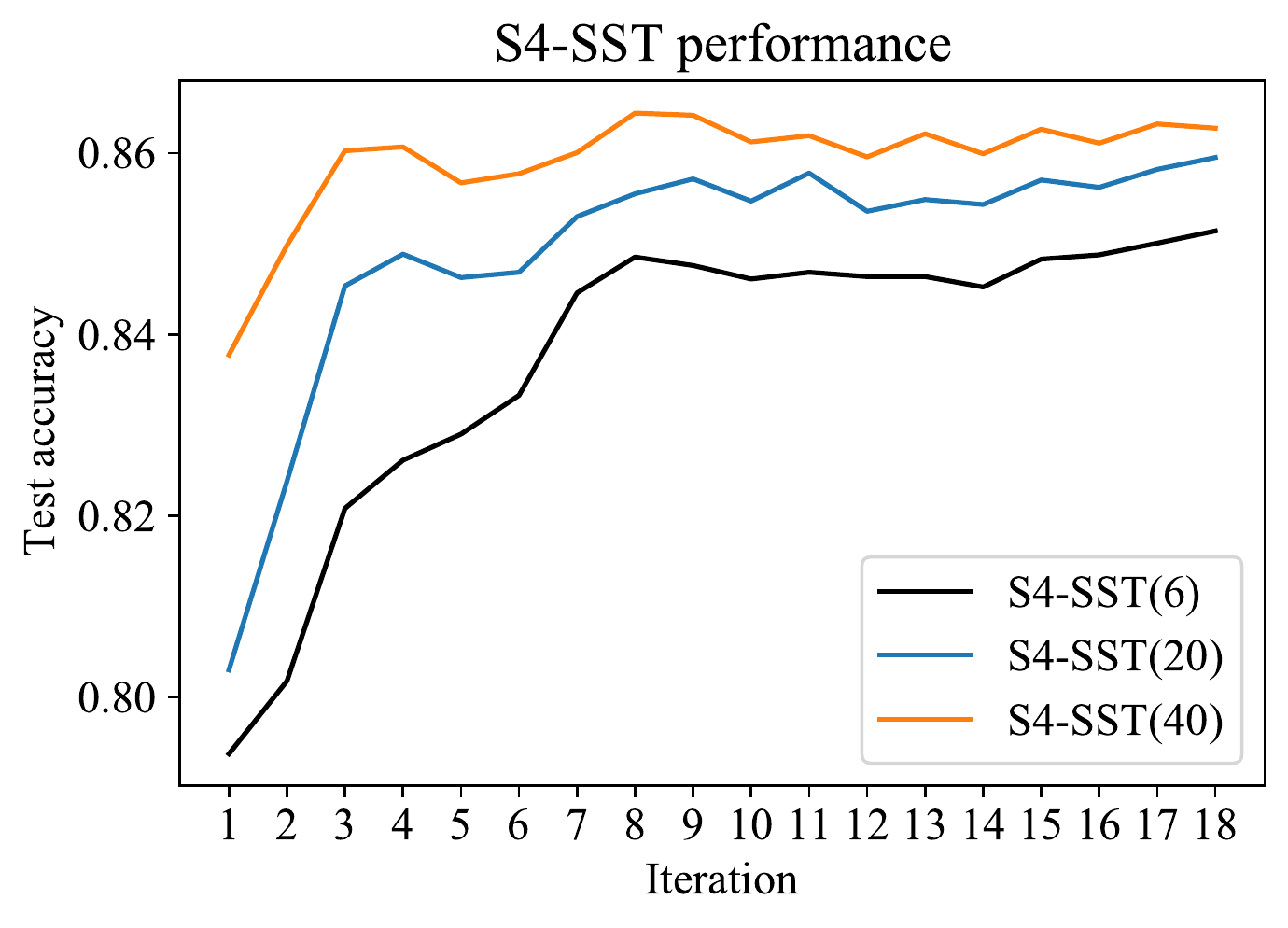} \\
    {\footnotesize (a)}
\end{tabular}
\begin{tabular}[b]{c}
    \includegraphics[width=0.45\textwidth]{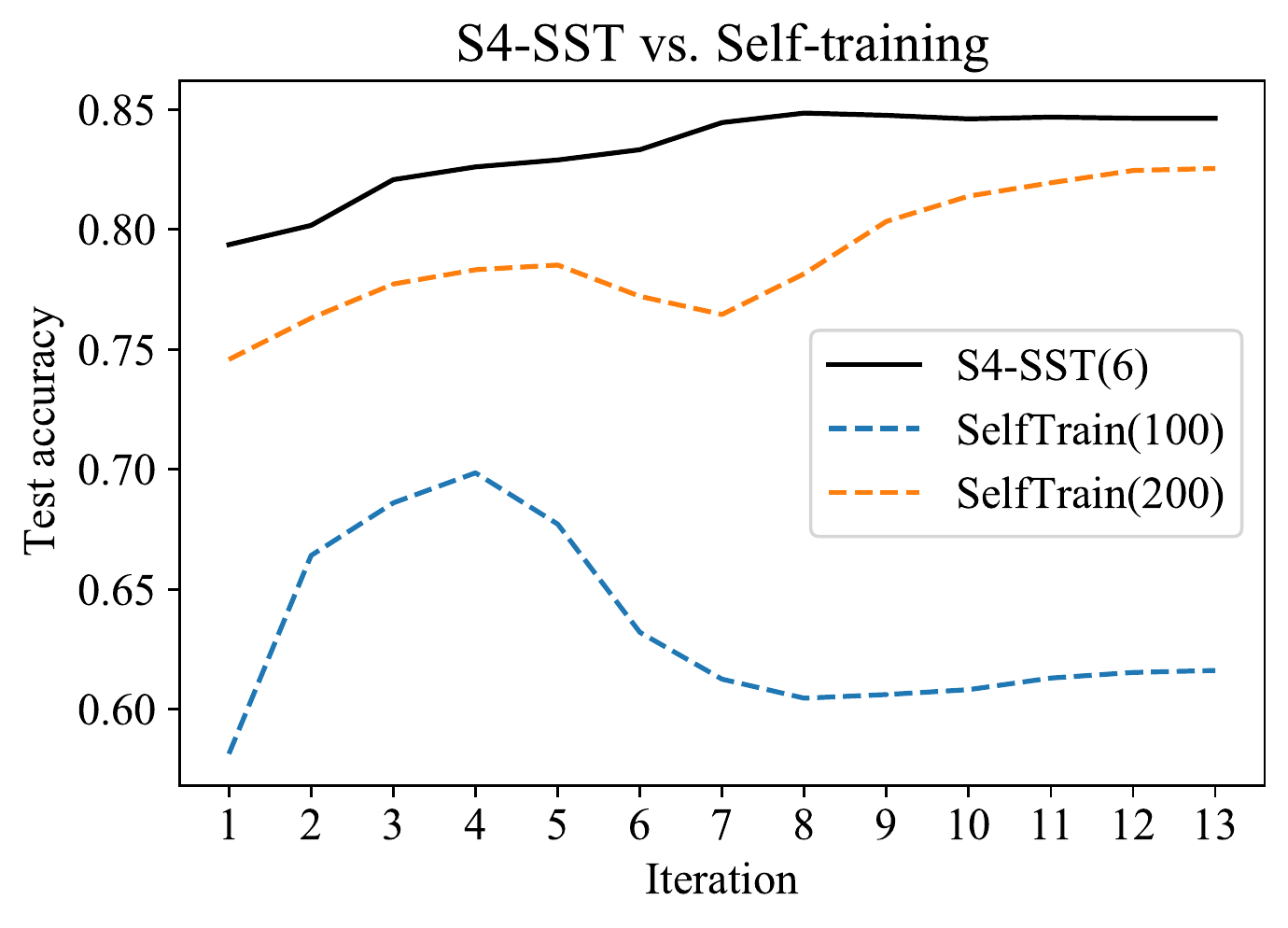} \\
    {\footnotesize (b)}
\end{tabular}
\begin{tabular}[b]{c}
    \includegraphics[width=0.45\textwidth]{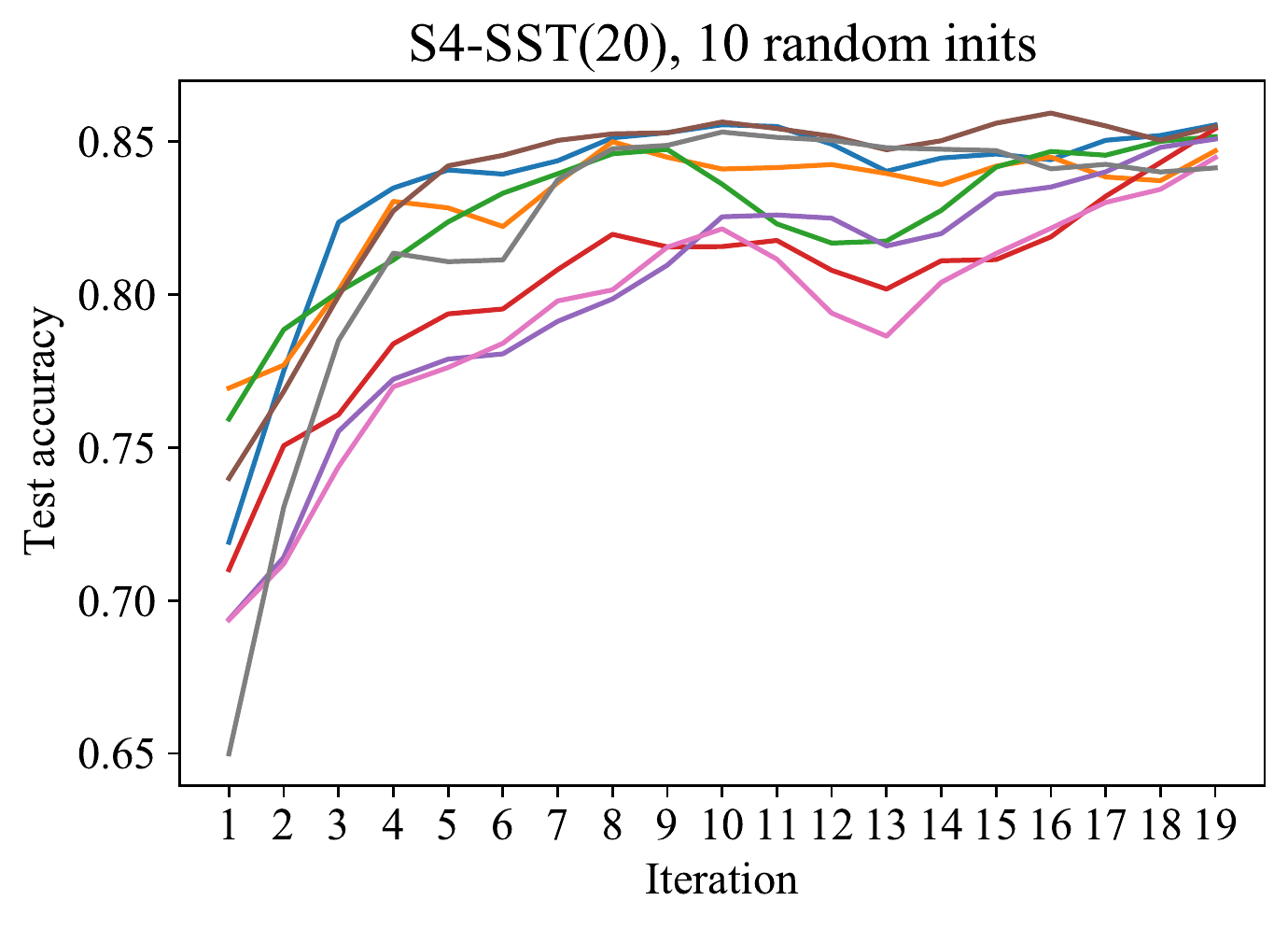} \\
    {\footnotesize (c)}
\end{tabular}
\begin{tabular}[b]{c}
      \resizebox{0.45\textwidth}{!}{
    \begin{tabular}[t]{cc|cc}
        \toprule
        Init. pos.      & Init. neg. & New pos. & New neg. \\
        \midrule
        refreshing &  lacks & superb &  terrible \\
        moving     &  waste & lovely     &  horrible \\
        caught     &  bland & touching    &  negative \\ 
        magic      &  stink & charming      &  boring \\
        captures   &   1    & delightful   &   2    \\
        \bottomrule
    \end{tabular}
    } \\ \\ \\
    {\footnotesize (d)}
\end{tabular}
\caption{S4-SST learning in IMDb.
(a) S4-SST learning curves with various numbers of initial self-supervised tokens. (b) comparison between the learning curves of S4-SST with just six self-supervised tokens and self-training with an order of magnitude more direct supervision (100/200 labeled examples). (c) multiple runs of S4-SST with 20 \emph{random} oracle self-supervised tokens (rather than the top oracle tokens). (d) pre-specified self-supervised tokens and proposed tokens in the first few iterations.
}
\label{fig:sst-best}
\end{figure}

Figure~\ref{fig:sst-best} (a) shows how S4-SST iterations improve the test accuracy of the learned neural network with different amounts of initial virtual evidence. 
Not surprisingly, with more initial self-supervision, the gain is less pronounced, but still significant. 
Figure~\ref{fig:sst-best} (b) compares the learning curves of S4-SST with those of self-training. 
Remarkably, with just \emph{six} self-supervised tokens, S4-SST not only attained substantial gain over the iterations, but also easily outperformed self-training despite the latter using an order of magnitude more label information (up to 200 labeled examples). 
This shows that S4 is much more effective in leveraging bounded human effort for supervision.

Figure~\ref{fig:sst-best} (c) shows 10 runs of S4-SST when it was initialized with 20 \emph{random} oracle tokens (10 per class), rather than the \emph{top} 20 tokens from the oracle. As expected, DPL's initial performance is worse than with the top oracle tokens.
However, over the iterations, S4-SST was able to recover even from particularly poor initial state, gaining up to 20 absolute accuracy points over DPL in the process. 
The final accuracy of S4-SST is 85.2 $\pm$ 0.9, compared to 71.5 $\pm$ 6.5 for DPL and 72.9 $\pm$ 6.7 for Snorkel, a mean improvement of more than 12 absolute accuracy points over both baselines. 
S4-SST's gains over DPL and Snorkel are statistically significant using a paired $t$-test (samples are paired when the algorithms have the same initial factors $I$) with $p=0.01$.
This indicates that S4-SST is robust to noise in the initial self-supervision.
Figure~\ref{fig:sst-best} (d) shows the initial self-supervised tokens and the ones learned by S4-SST in the first few iterations. We can see that S4-SST is able to learn highly relevant self-supervision tokens.

S4 has similar gains over DPL, Snorkel, and self-training on the Stanford dataset. See Table~\ref{tbl:sys} (b).
The Stanford dataset is much more challenging and for a long while, it was hard to exceed 80\% test accuracy \cite{socher2013recursive}.  Interestingly, S4-SST was able to surpass this milestone using just 20 initial self-supervision tokens. As in IMDb, the only randomness is in the initialization of $\Psi$, which is negligible. 

For IMDb, as can be seen above, even with simple token-based self-supervision, S4-SST already performed extremely well. So we focused our investigation of joint-inference factors on the more challenging Stanford dataset (S4-SST + J). 
While S4-SST already performed very well with token-based self-supervision, by incorporating joint-inference factors, it could still gain up to 3 absolute accuracy points.
An example learned joint-inference factor is between the sentence pair:  
{\em This is no ``Waterboy!"} and {\em ``It manages to accomplish what few sequels can---it equals the original and in some ways even better"}.
Note that Waterboy is widely considered a bad movie, hence the first sentence expresses a positive sentiment, just like the second.
It is remarkable that S4 can automatically induce such factors with complex and subtle semantics from small initial self-supervision. 

The Stanford results also demonstrated that active learning could play a bigger role in more challenging scenarios. With limited initial self-supervision ($|I|=6$), the full S4 system (S4+J (T=20)) gained 8 absolute points over S4-SST and 5 absolute points over S4-SST+J. With sufficient initial self-supervision and joint-inference, however, active learning was actually slightly detrimental ($|I|=20,40$).

\begin{table}
\centering
\begin{tabular}[b]{c}
      \resizebox{0.45\textwidth}{!}{
        \begin{tabular}{lcccc}
            \toprule
            & $|I|=6$ & $|I|=10$ & $|I|=20$ & $|I|=40$\\
            \midrule
            $T=5$ & 79.0 & 83.8 & 84.0 & 85.9  \\
            $T=10$ & 79.5 & 83.7 & 85.0 & 86.0\\
            $T=20$ & 82.7 & 84.5 & 85.8 & 86.4\\
            \bottomrule
        \end{tabular}%
        }\\
        {\footnotesize (a) IMDb}
\end{tabular}
\begin{tabular}[b]{c}
      \resizebox{0.45\textwidth}{!}{
        \begin{tabular}{lcccc}
            \toprule
            & $|I|=6$ & $|I|=10$ & $|I|=20$ & $|I|=40$\\
            \midrule
            $T=5$  & 71.4  & 77.5 & 82.5 & 83.9 \\
            $T=10$ & 72.3 & 77.4 & 82.5 & 83.7 \\
            $T=20$ & 77.1 & 80.8 & 83.4 & 84.0 \\
            \bottomrule
        \end{tabular}  
        }\\
        {\footnotesize (b) Stanford}
\end{tabular}
        \caption{S4-FAL test results (no SST steps)}\label{tbl:fal}
\end{table}

Finally, we evaluate S4-FAL, which conducts active learning but not structure learning. See Table~\ref{tbl:fal}. As expected, performance improved with larger initial self-supervision ($I$) and human query budget ($T$). Active learning helps the most when initial self-supervision is limited. Compared to S4 with structure learning, however, active learning alone is less effective. For example, without requiring any human queries, S4-SST outperformed S4-FAL on both IMDB and Stanford even when the latter was allowed up to $T=20$ human queries.

\begin{table}[!htb]
\centering
\subfloat[][IMDb]{
      \resizebox{0.45\textwidth}{!}{
        \begin{tabular}{ccc}
        %\centering
            \toprule
            $|I|$ & Entropy-based  & Attention-based\\
            \midrule
            6 & 82.1 & 85.5 \\
            20 & 84.9 & 86.4 \\
            40 & 85.5 & 86.6\\
            \bottomrule
        \end{tabular}
        }
        }
\hfill
\subfloat[][Stanford]{        
      \resizebox{0.45\textwidth}{!}{
        \begin{tabular}{ccc}
         %\centering
            \toprule
            $|I|$ & Entropy-based  & Attention-based\\
            \midrule
            6 & 77.2 & 73.0\\
            %10 & 84.5\\
            20 & 85.1 & 83.3 \\
            40 & 85.7 & 84.9 \\
            %80 & 86.2\\
            \bottomrule
        \end{tabular}
        }
        }
    \caption{Entropy-based scoring ($S_\text{entropy}$) perform comparably as attention-based scoring ($S_\text{token}$).}\label{tbl:entropy-attn}
\end{table}

\begin{table}[!htb]
    \centering
    \resizebox{0.45\textwidth}{!}{
        \begin{tabular}{lcc}
            \toprule
            Algorithm   & Sup. size   & Test acc (\%) \\
            \midrule
            BoW & 140k &  71.2   \\
            DNN & 140k &  79.8  \\
            \midrule
            Self-training & 50 & 38.4 \\
            Snorkel & 50  & 37.2 \\        
            DPL & 50  & 41.8  \\
            S4-SST & 50 & 49.1  \\            
            \midrule
            Self-training & 100 & 38.2  \\
            Snorkel & 100  & 36.5\\        
            %Snorkel + SST factors & 100 & 44.2\\
            DPL & 100  & 41.7  \\
            S4-SST & 100 & 52.3  \\
            \bottomrule
        \end{tabular}
        }
        \caption{Comparison of test accuracy on Yahoo! Answers.}\label{tbl:yahoo}
\end{table}

Table \ref{tbl:entropy-attn} compares S4-SST test accuracy using entropy-based scoring ($S_\text{entropy}$) and attention-based scoring ($S_\text{token}$). Entropy-based scoring slightly outperforms attention-based scoring on Stanford Sentiment and slightly trails on IMDb. Overall, the two perform comparably but entropy-based scoring is more generally applicable.

S4-SST obtains similarly substantial gains on Yahoo using token-based factors and attention scoring. See Table \ref{tbl:yahoo}. We focused our experiments on a fixed 10\% of the training set due to its very large size (1.4 million examples). There are 10 classes, so initial self-supervision size of 50 (100) represents 5 (10) initial tokens per class (for S4, DPL, and Snorkel), or 5 (10) labeled examples per class (for self-training). 
Compared to binary sentiment analysis in IMDb and Stanford, Yahoo represents a much more challenging dataset, with ten classes and larger input text for each instance. The linguistic phenomena are much more diverse, and therefore neither Snorkel or DPL performed much better than self-training, as token-based self-supervision confers not much more information than a labeled example. However, S4-SST is still able to attain substantial improvement over the initial self-supervision. E.g., with initial supervision size of 100, S4-SST gained about 11 absolute accuracy points over DPL, and 16 absolute points over Snorkel. 
Additionally, S4-SST is able to better utilize the new factors than Snorkel. If we run Snorkel using the same initial factors as S4-SST and also add the new factors proposed by S4-SST in each iteration, the accuracy improved from 36.5 to 44.2, but still trailed S4-SST (52.3) by 8 absolute points. 
Interestingly, both DPL and Snorkel perform better on Yahoo with \emph{fewer} initial factors, at 5 per class, suggesting they are sensitive to noise in the less reliable initial self-supervision. By contrast, S4-SST is more noise-tolerant and benefits from additional initial supervision.

%%%%%%%%%%%%%%%%%%%%%%
\eat{
\subsection{Details and additional experiments}
\label{sec:alt-scores}
\subsubsection{Structured Self-Training Convergence}
In Algorithm \ref{alg:S4}, structured self-training (SST) iterations are repeated until convergence in the while loop. Here we elaborate on the convergence criterion described in the main text (Self-Supervised Self-Supervision Section). Intuitively, convergence occurs when the expected latent labels change little despite the addition of new self-supervision from SST.

Formally, consider the set $\Delta = \{i  | \argmax_{y_i} E_{\Phi^{t-1}(X,Y)}[y_i] \ne \argmax_{y_i} E_{\Phi^{t}(X,Y)}[y_i]\}$. This is the set of instances for which the labels based on self-supervision alone (\emph{excluding} the neural network prediction module $\Psi$) have changed between subsequent iterations. We stop the SST iterations once $|\Delta|/N < \alpha$ for some small $\alpha$, as we don't expect there will be much change afterwards. We used $\alpha=1\%$, which worked well in preliminary experiments, and performed no further tuning.

\subsubsection{Count Normalization in Score Functions}
All the score functions in $\tt PropSST$ and $\tt PropFAL$ normalize the score using the feature count (e.g., $C_t$, $C_b$). In practice, if we apply this to all features, we may inadvertently promote rare features. Thus \cite{druck2009active} additionally multiplied by the logarithm of the count (i.e., they normalize using $\frac{\log C_t}{C_t}$). By contrast, we found it preferable to use standard count normalization, but simply skip the rare features. In all of our experiments, we only consider the top 2.5\% most frequent features. We found that this worked well in preliminary experiments on the held-out data in IMDb, and our results are not sensitive to this value.

\subsubsection{Additional results and hyperparameters}
\smallpar{Yahoo Dataset}
We provide additional results for S4-SST on Yahoo using token-based factors and attention scoring. See Table \ref{tbl:yahoo}. We focused our experiments on a fixed 10\% of the training set due to its very large size (1.4 million examples). 
There are 10 classes, so initial self-supervision size of 50 (100) represents 5 (10) initial tokens per class (for S4, DPL, and Snorkel), or 5 (10) labeled examples per class (for self-training). 
Compared to binary sentiment analysis in IMDb and Stanford, Yahoo represents a much more challenging dataset, with ten classes and larger input text for each instance. The linguistic phenomena are much more diverse, and therefore neither Snorkel or DPL performed much better than self-training, as a token-based self-supervision confers not much more information than a labeled example. However, S4-SST is still able to attain substantial improvement over the initial self-supervision. E.g., with initial supervision size of 100, S4-SST gained about 11 absolute accuracy points over DPL, and 16 absolute points over Snorkel.  
Additionally, S4-SST is able to better utilize the new factors than Snorkel. If we run Snorkel using the same initial factors as S4-SST and also add the new factors proposed by S4-SST in each iteration, the accuracy improved from 36.5 to 44.2, but still trailed S4-SST (52.3) by 8 absolute points. 
Interestingly, both DPL and Snorkel perform better on Yahoo with \emph{fewer} initial factors, at 5 per class, suggesting they are sensitive to noise in the less reliable initial self-supervision. By contrast, S4-SST is more noise-tolerant and benefits from additional initial supervision.  

}

\smallsection{Conclusions}

In this chapter, we present deep probabilistic logic (DPL) as a unifying framework for task-specific self-supervised learning, by composing probabilistic logic with deep learning. We further present self-supervised self-supervision (S4), which extends DPL with structure learning and active learning capabilities.  Experiments on biomedical machine reading and text classification tasks show that DPL enables effective combination of disparate task-specific self-supervision strategies, while S4 can further improve application performance by proposing additional self-supervision. Overall, this enables the most effective use of human expert bandwidth by focusing on identifying the most salient self-supervision for initialization and verifying proposed self-supervision. While we focus on NLP tasks in this chapter, our methods are general and can potentially be applied to other domains. 
Future directions include: combining DPL with deep generative models and probabilistic programming methods; further investigation in combining structure learning and active learning in S4; exploring alternative optimization strategies and more sophisticated self-supervision classes and proposal algorithms; applications to other domains.

\bibliographystyle{tfnlm}
\bibliography{reference}

\end{document}